\documentclass[10pt,twocolumn,letterpaper]{article}

\usepackage{cvpr}              

\usepackage{graphicx}
\usepackage{amsmath}
\usepackage{amssymb}
\usepackage{booktabs}
\usepackage{microtype}
\usepackage{soul}
\usepackage{sidecap}
\usepackage{enumitem}

\newenvironment{tight_itemize}{
\begin{itemize}[leftmargin=15pt,nosep]
  \setlength{\topsep}{0pt}
  \setlength{\itemsep}{0pt}
  \setlength{\parskip}{0pt}
  \setlength{\parsep}{0pt}
}{\end{itemize}}

\newcommand{\Ours}{IRISformer}
\newcommand{\Oursb}{\textbf{IRISformer}}

\usepackage[pagebackref,breaklinks,colorlinks]{hyperref}

\usepackage[capitalize]{cleveref}
\crefname{section}{Sec.}{Secs.}
\Crefname{section}{Section}{Sections}
\Crefname{table}{Table}{Tables}
\crefname{table}{Tab.}{Tabs.}

\def\cvprPaperID{6661} 
\def\confName{CVPR}
\def\confYear{2022}

\begin{document}

\title{IRISformer: Dense Vision Transformers for Single-Image Inverse Rendering in Indoor Scenes}

\author{
{Rui Zhu$^{1}$}\quad
{Zhengqin Li$^{1}$}\quad
{Janarbek Matai$^{2}$}\quad
{Fatih Porikli$^{2}$}\quad
{Manmohan Chandraker$^{1}$}\quad
\\[2mm]
{$^{1}$UC San Diego }\quad
{$^{2}$Qualcomm AI Research}
\\
{\tt\small \{rzhu,zhl378,mkchandraker\}@eng.ucsd.edu}\quad
{\tt\small \{jmatai,fporikli\}@qti.qualcomm.com}
}

\maketitle

\begin{abstract}


Indoor scenes exhibit significant appearance variations due to myriad interactions between arbitrarily diverse object shapes, spatially-changing materials, and complex lighting. Shadows, highlights, and inter-reflections caused by visible and invisible light sources require reasoning about long-range interactions for  inverse rendering, which seeks to recover the components of image formation, namely, shape, material, and lighting. 
In this work, our intuition is that the long-range attention learned by transformer architectures is ideally suited to solve longstanding challenges in single-image inverse rendering. We demonstrate with a specific instantiation of a dense vision transformer, \Ours{}, that excels at both single-task and multi-task reasoning required for inverse rendering. Specifically, we propose a transformer architecture to simultaneously estimate depths, normals, spatially-varying albedo, roughness and lighting from a single image of an indoor scene. Our extensive evaluations on benchmark datasets demonstrate state-of-the-art results on each of the above tasks, enabling applications like object insertion and material editing in a single unconstrained real image, with greater photorealism than prior works. Code and data are publicly released.\footnote{\url{https://github.com/ViLab-UCSD/IRISformer}}

\end{abstract}

\section{Introduction}
\label{sec:intro}

\begin{figure}
\centering
\includegraphics[width=\columnwidth]{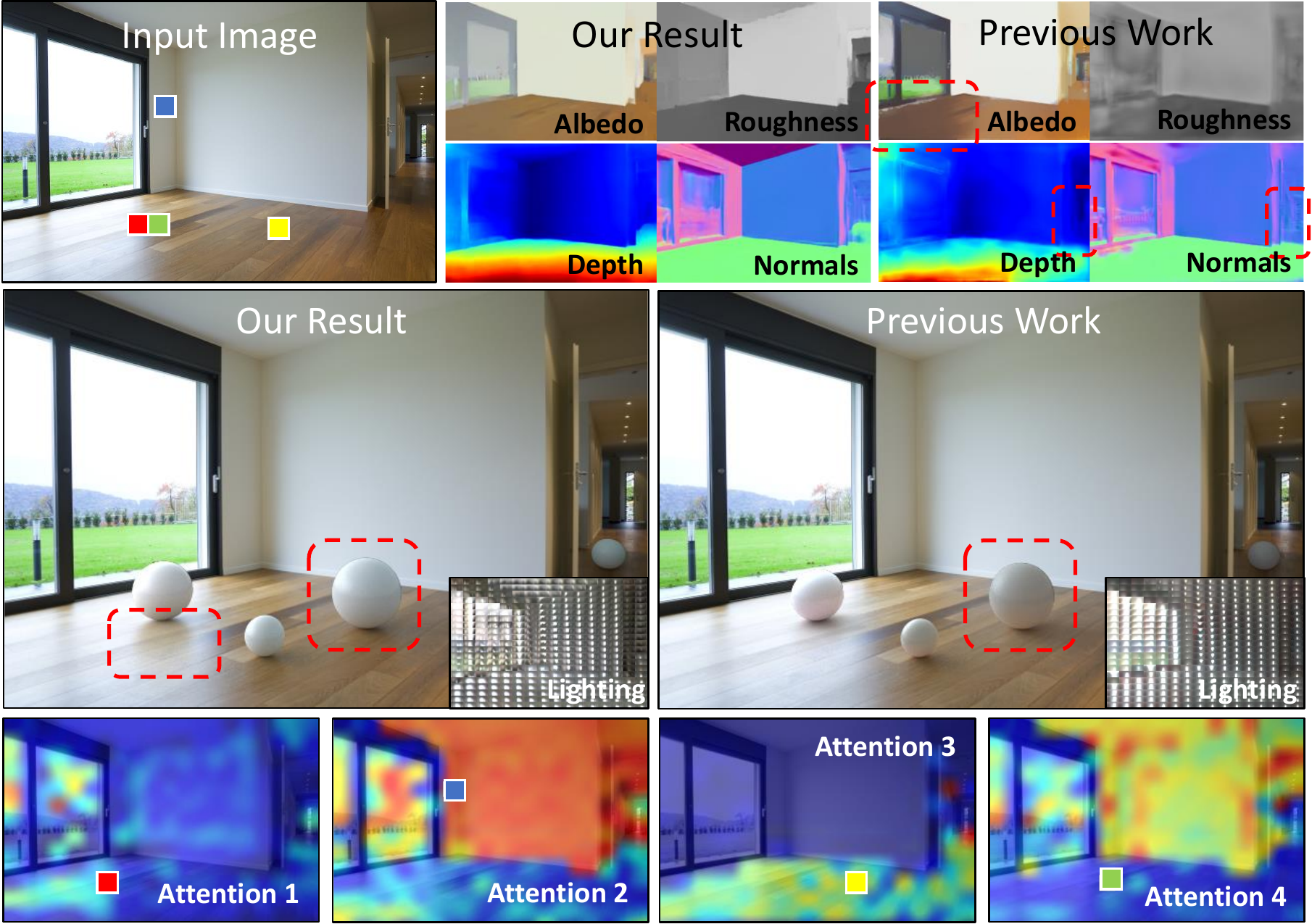}
\caption{Given a single real world image, \Ours{} simultaneously infers material (albedo and roughness), geometry (depth and normals), and spatially-varying lighting of the scene. The estimation enables virtual object insertion where we demonstrate high-quality photorealistic renderings in challenging lighting conditions compared to previous work~\cite{li2020inverse}. The learned attention is also visualized for selected patches, indicating benefits of global attention to reason about distant interactions (see text for details).}
\label{fig:teaser}
\end{figure}

Inverse rendering has long been of great interest to the computer vision community owing to its promise to decompose a scene into the intrinsic factors of shape, complex spatially-varying lighting, and material, thereby enabling downstream tasks of virtual object insertion, material editing, and relighting. The problem is particularly challenging for indoor scenes, where complex appearances stem from multiple interactions among the above intrinsic factors, such as shadows, specularities, and interreflections. 

Recent advances in inverse rendering has led to the emergence of numerous works undertaking either some specific aspects of this challenge (geometry~\cite{eigen2015predicting, liu2019planercnn}, albedo~\cite{li2018cgintrinsics, sengupta2019nir, bell2014iiw}, lighting~\cite{gardner2017learning, srinivasan2020lighthouse, legendre2019deeplight, gardner2019deeppara}), or joint estimation~\cite{barron2013intrinsic, sengupta2019nir, li2020inverse, wang2021learning}. However, the task of scene decomposition can be extremely ill-posed due to the inherent ambiguity between complex lighting, geometry, and material which jointly govern image formation in indoor scenes. For example, high-intensity pixel values can be explained by either specular or light-colored material, particular local geometry, bright lighting, or by a combination of all those factors. The problem is especially severe with only a single image as input, where prior knowledge is necessary to disambiguate among all possible intrinsic decompositions that explain the image. Classical methods leverage strong heuristic priors in an optimization objective~\cite{barron2013intrinsic, barron2014shape}, which may not always hold for real world scenes with complex geometry or lighting conditions.

The widespread use of convolutional neural networks (CNN) and large-scale datasets for scene decomposition~\cite{bell2014iiw,silberman2012nyu,sengupta2019nir,li2021openrooms} promotes supervised training of end-to-end multi-task models~\cite{sengupta2019nir, li2020inverse} for joint estimation. CNN-based models have demonstrated impressive progress on inverse rendering of real world images \cite{li2020inverse,sengupta2019nir,wang2021learning, srinivasan2020lighthouse}. Nonetheless, receptive fields in CNN architectures remain largely local throughout the consecutive layers, limiting the ability to capture long-range interactions between scene elements. As shown in Fig.~\ref{fig:teaser}, CNN-based approaches fail to handle scenes where strong shadows or highlights abound due to complex light transport. This indicates that long-range dependencies across the image space must be exploited to provide globally coherent predictions in inverse rendering. Recently, vision transformers~\cite{dosovitskiy2020vit, yuan2021tokens} (ViT) have emerged for multiple computer vision tasks, benefiting from global reasoning via spatial attention mechanisms. In particular, dense vision transformers~\cite{ranftl2021dpt, liu2021swin, wang2021pyramid} are well-suited for dense prediction, which we posit can benefit inverse rendering.

In this paper, we propose to leverage vision transformers to better account for complex light transport in inverse rendering. Consider Fig.~\ref{fig:teaser} as an example, where we compare our proposed transformer-based approach, \ie \Oursb{} (Trans\textit{former} for \textit{I}nverse \textit{R}endering in \textit{I}ndoor \textit{S}cenes), with a CNN-based prior state-of-the-art \cite{li2020inverse}. Note the improvement in material consistency and geometry of the floor where complex lighting governs appearance; as a result of which, the leftmost sphere is properly reflected on the floor. Additionally, \Ours{} better captures global ambient lighting so that the third sphere from left is better illuminated.
We also visualize the heatmaps of four patch locations shown by colored squares from selected transformer layers and heads (warmer colors indicate higher attention).
By attending to large \textit{global} regions with semantic meaning, the transformer can better disambiguate geometry material and lighting (yellow). Long-range interactions among such regions can help reason about inter-reflections (green), directional highlights (red), or shadows (blue). as well as the long-range attention to/within those homogeneous regions, the model manages to better resolve the  albedo-lighting ambiguity, and making more consistent estimations. 

We demonstrate that by the insightful design of single-task and multi-task models for inverse rendering with dense vision transformers, we can achieve state-of-the-art, high-quality, and globally coherent BRDF, geometry, and lighting prediction. In addition, downstream tasks like object insertion and material editing greatly benefit from our improvements, especially in scenarios of complex highlights or shadows. We achieve state-of-the-art results on all sub-tasks on real world datasets of IIW~\cite{bell2014iiw} and NYUv2~\cite{silberman2012nyu}, as well as object insertion tasks compared to prior works.

Our contributions are threefold. (1) We propose the first dense vision transformer-based framework for inverse rendering in a multi-task setting. (2) We demonstrate that appropriate design choices lead to better handling of global interactions between scene components, leading to better disambiguation of shape, material and lighting. (3) We demonstrate state-of-the-art results on all tasks and enable high-quality applications in augmented reality.

\begin{figure*}
\centering
\includegraphics[width=\textwidth]{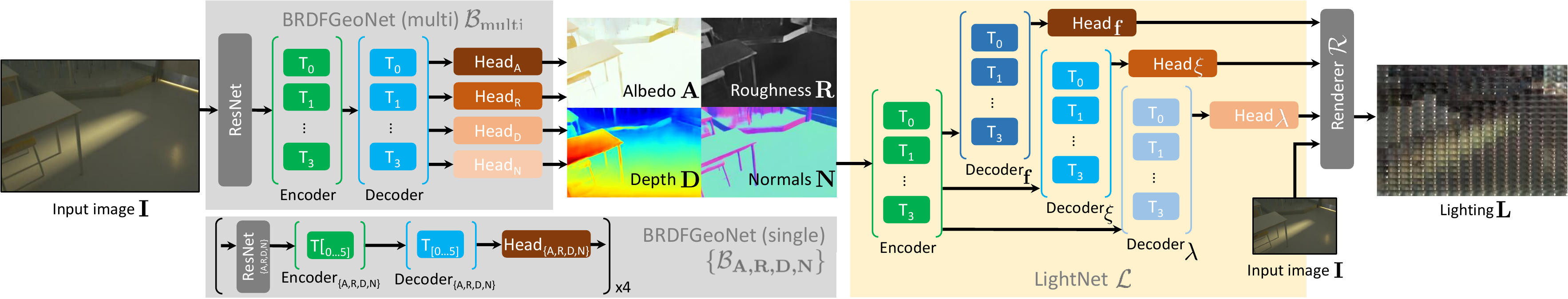}
\vspace{-0.4cm}
\caption{Overview of \Ours{}. For \textbf{BRDFGeoNet}, we illustrate the multi-task setting in the upper gray block, while the single-task case (lower gray block) has 4 independent copies of DPT with different design of output heads.}
\vspace{-0.2cm}
\label{fig:pipeline}
\end{figure*}

\section{Related Work}
\label{sec:related}

\noindent\textbf{Inverse rendering for indoor scenes.}
Several prior works study the interaction of shape, material and light to estimate shape-from-shading~\cite{horn1989shapefromshading,zhang1999shapefromshading}, intrinsic image decomposition~\cite{li2018cgintrinsics, grosse2009ground}, material properties \cite{aittala2016reflectance,deschaintre2018single,li2018material,li2018learning,sang2020}, or illumination~\cite{barron2014shape,gardner2017learning,gardner2019parametric,zhan2021gmlight}, while inverse rendering seeks to estimate all those factors simultaneously~\cite{Marschner1998InverseRF}. Classical methods for inverse rendering are typically posed as energy minimization with heuristic priors, for example, SIRFS~\cite{barron2013intrinsic,barron2014shape} where intrinsic properties are jointly optimized with a statistical cost function. Despite early success, such models usually do not generalize well to real images with diverse appearances. With advances in deep learning, CNN-based methods have been developed to learn a generalizable model in a data-driven fashion. The recent method of NIR~\cite{sengupta2019nir} is pre-trained with weak labels and finetuned on real images with re-rendering loss. Methods like Lighthouse~\cite{srinivasan2020lighthouse} learn a volumetric lighting representation using stereo inputs, while Wang~\etal~\cite{wang2021learning} do so with a single image. Li~\etal~\cite{li2020inverse,li2021openrooms} design physically-based representations and rendering layers to estimate shape, SVBRDF, and spatially-varying per-pixel lighting from a single image. However, the aforementioned methods utilize convolutional neural networks, which feature a limited receptive field and lack an explicit attention mechanism to reason about long-range dependencies in image space, which can be crucial for estimating global properties of light transport and its interaction with material and shape.

\vspace{0.2cm}
\noindent\textbf{Datasets for inverse rendering.}
Synthetic datasets~\cite{song2017suncg, li2021openrooms, roberts2020hypersim} are commonly used to provide ground truth for most modalities, including scene geometry, material, and lighting, and suitable for training inverse rendering models in supervised fashion~\cite{li2020inverse}, while achieving good generalizability to real world datasets. Models trained on synthetic datasets are shown to further improve on real world images via finetuning with either weak supervision~\cite{bell2014iiw}, full supervision on a subset of modalities~\cite{silberman2012nyu}, or with re-rendering losses~\cite{sengupta2019nir}. In this work, we train our transformer models on the OpenRooms dataset \cite{li2021openrooms} and obtain state-of-the-art results by finetuning on IIW~\cite{bell2014iiw} and NYUv2~\cite{silberman2012nyu} real world datasets.

\vspace{0.2cm}
\noindent\textbf{Vision transformer.}
Convolutional neural networks (CNN) have long been the architecture of choice as building blocks for dense prediction with deep learning, as both backbone for feature extraction ~\cite{he2016deep,chen2017deeplab} or as decoders~\cite{newell2016stacked,li2020inverse}. However, several drawbacks inherent to CNNs make them sub-optimal for tasks that require reasoning over long-range dependencies over the image space, despite multiple measures that have been proposed to mitigate those issues, including dilated convolutions~\cite{chen2017deeplab,yu2015dilated}, skip connections ~\cite{he2016deep,ronneberger2015u} and self-attention~\cite{zhang2019self}. The recently proposed Vision Transformer (ViT)~\cite{dosovitskiy2020vit} has enabled feature extraction with global attention over image space with elegant design and better interpretability while achieving superior performance compared to CNNs on multiple vision tasks. Several works~\cite{wang2021pyramid,yuan2021tokens,xu2021yogo} have extended ViT to dense prediction tasks, including DPT~\cite{ranftl2021dpt}, Swin Transformer~\cite{liu2021swin}, etc. Additional efforts have been made to utilize transformers in a multi-task or multi-object setting, such as UniT~\cite{hu2021unit} that follows an encoder-decoder design scheme and utilizes a task-specific query embedding to learn a unified decoding feature space for all tasks. In contrast to those works, we propose single-task and multi-task transformers for dense prediction tasks in inverse rendering, where material and lighting estimation using transformers have previously not been studied.

\section{Method}
\label{sec:method}

\noindent\textbf{Notation.}
Vectors are represented with a lower-case bold font (\eg $\mathbf{x}$). Matrices are in upper-case bold (\eg $\mathbf{X}$) while scalars are in regular font (\eg $x$ or $X$). 
Variables with hat, \eg $\mathbf{\hat X}$, are the estimation of the corresponding entity $\mathbf{X}$. For denoting the $l^{th}$ sample in a set (\eg images, shapes), we use subscripts (\eg $\mathbf{X}_{l}$). Uppercase calligraphic symbols (\eg$\mathcal{X}$) denote functions. 

\subsection{Scene Representation and Loss Functions}
\noindent\textbf{Geometry and spatially-varying material.} 
In \Ours{}, we account for only the scene elements within the camera frustum. For an $h \times w$ image, we represent per-pixel geometry with a depth map $\mathbf{D} \in \mathbb{R}^{h\times w}$ and normals $\mathbf{N} \in \mathbb{R}^{h\times w \times 3}$. We represent material as a microfacet spatially-varying BRDF model (SVBRDF)~\cite{karis2013svbrdf}, with albedo $\mathbf{A}$ and roughness $\mathbf{R}$ maps of sizes $h \times w \times 3$ and $h \times w$ respectively. Specifically, given a single RGB input image $\mathbf{I}$, we seek to learn a function \textbf{BRDFGeoNet} denoted as $\mathcal{B}$ to jointly estimate the above properties: $\{\mathbf{\hat D}, \mathbf{\hat N}, \mathbf{\hat A}, \mathbf{\hat R}\} = \mathcal{B}(\mathbf{I})$.
Given ground truths, we use scale-invariant L2 loss~\cite{li2020inverse, ranftl2019midas} for albedo and depth ($\log$ space) and an L2 loss for roughness and normals.

\vspace{0.2cm}
\noindent\textbf{Spatially-varying lighting.}
For lighting, we follow Li~\etal~\cite{li2020inverse,li2021openrooms} to adopt per-pixel image-space environment maps $\mathcal{G}$ of $16 \times 32$ pixels to represent the incident irradiance, parameterized by $K=12$ Spherical Gaussian (SG) lobes $\{\xi_k, \lambda_k, \textbf{f}_k\}_{k=1}^{K}$. Here $\xi_k \in \mathbb{S}^2$ is the center orientation on the unit sphere, $\textbf{f}_k \in \mathbb{R}^3$ is the intensity and $\lambda_k \in \mathbb{R}$ is the bandwidth. Given each set of lighting parameters and an orientation outward from one spatial location $\eta$ in 3D, we have
\begin{equation}
    \mathcal{G}(\eta) = \sum_{k=1}^{K} \textbf{f}_k e^{\lambda (1 - \eta \xi)}:\mathbb{S}^2 \rightarrow \mathbb{R}^3.
\end{equation}
For a collection of all $h\times w$ pixel locations and all $16\times 32$ outgoing directions from each surface point in 3D, we arrive at a lighting map $\mathbf{L} \in \mathbb{R}^{h \times w \times 16 \times 32 \times 3}$. Then, using either a single image as the sole input or combining it with predictions of geometry and material, a \textbf{LightNet} denoted as $\mathcal{L}$ can be learned: $\mathbf{\hat L} = \mathcal{L}(\mathbf{I})$ or $\mathbf{\hat L} = \mathcal{L}(\mathbf{I}, \mathbf{\hat D}, \mathbf{\hat N}, \mathbf{\hat A}, \mathbf{\hat R})$.
%
Given estimates of geometry, material and lighting, a per-pixel physically-based differentiable renderer \textbf{RenderNet}~\cite{li2020inverse} denoted as $\mathcal{R}$ can re-render the input image: $\mathbf{\hat I} = \mathcal{R}(\mathbf{\hat D}, \mathbf{\hat N}, \mathbf{\hat A}, \mathbf{\hat R}, \mathbf{\hat L})$.
%

To supervise the training of lighting, assuming dense ground truth can be acquired from synthetic datasets, we impose a scale-invariant L2 reconstruction loss on the lighting map $\mathbf{\hat L}$ (in $\log$ space) and a scale-invariant L2 re-rendering loss on the re-rendered image $\mathbf{\hat I}$. The final loss $L_{\textrm{all}}$ is a weighted combination of losses on all estimations: 
\begin{equation}
    L_{\textrm{all}} = \lambda_{\textbf{A}} L_{\textbf{A}} + \lambda_{\textbf{R}} L_{\textbf{R}} + \lambda_{\textbf{D}} L_{\textbf{D}} + \lambda_{\textbf{N}} L_{\textbf{N}} + \lambda_{\textbf{L}} L_{\textbf{L}} + \lambda_{\textbf{I}} L_{\textbf{I}}.
    \label{equ:loss}
\end{equation}

\begin{figure*}
\centering
\includegraphics[width=0.95\textwidth]{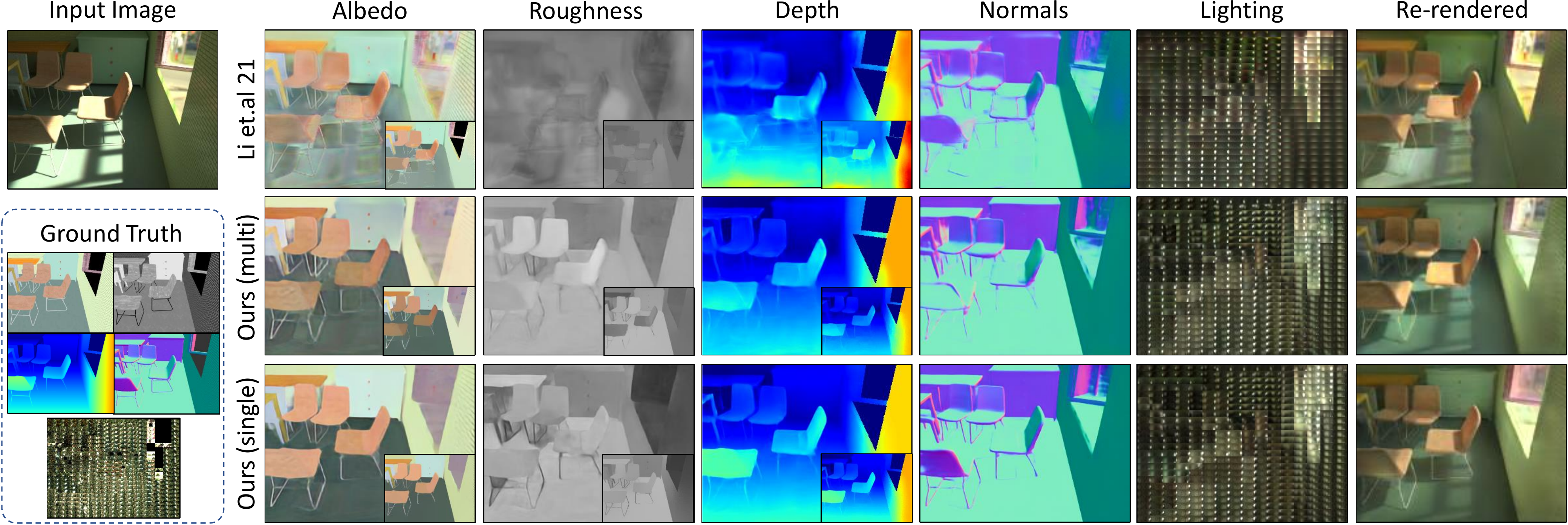}
\vspace{-0.2cm}
\caption{BRDF, geometry and lighting estimation on OpenRooms. Small insets (best viewed when enlarged in PDF version) are estimations processed with bilateral solvers (BS). More results can be found in the supplementary material.}
\vspace{-0.4cm}
\label{fig:or_results}
\end{figure*}

\subsection{Dense Vision Transformer}
Dense Vision Transformer (DPT)~\cite{ranftl2021dpt} is a generic architecture for dense prediction utilizing vision transformers~\cite{dosovitskiy2020vit} in place of CNNs as a backbone. An image $\mathbf{I}$ is first divided into an $h/p \times w/p$ grid of non-overlapping patches of size $p\times p$, and subsequently DPT tokenizes each patch into a vector of dimension $D$ with a shadow CNN (DPT-large or DPT-base) or ResNet (DPT-hybrid). The result is a set of tokens $\textbf{t}^0 = \{\textbf{t}^0_1, \hdots, \textbf{t}^0_{N_p}\}$ where $N_p = hw/p^2$, $\textbf{t}^0_n \in \mathbb{R}^D, n=[1, \dots N_p]$. A cascade of $M$ transformer layers then transforms the set of vectors with self-attention~\cite{vaswani2017attention} into $\textbf{t}^M$, and a re-assembling operation followed by a convolutional decoder transforms the tokens back to 2D space, resulting in a 2D dense feature map. A customized convolutional head is attached to yield the final prediction from the feature map, based on the specific prediction task.

\subsection{Single-task Network Design}
Due to the modular design of DPT and the variation of size and capacity among different DPT variants~\cite{ranftl2021dpt}, we consider a few design choices for using DPT modules to build \textbf{BRDFGeoNet} and \textbf{LightNet} in both single-task and multi-task settings.

The full design of our pipeline can be found in Fig.~\ref{fig:pipeline}. 
In single-task setting, we seek to maximize the performance of each task by using an independent DPT for each of depth, normal, albedo, roughness, and lighting. This effectively results in \textbf{BRDFGeoNet} of 4 DPTs $\{\mathcal{B}_A, \mathcal{B}_R, \mathcal{B}_D, \mathcal{B}_N\}$ to independent infer each of the modality: $\mathbf{D} = \mathcal{B}_D(\mathbf{I})$, $\mathbf{N} = \mathcal{B}_N(\mathbf{I})$, $\mathbf{A} = \mathcal{B}_A(\mathbf{I})$, $\mathbf{R} = \mathcal{B}_R(\mathbf{I})$. For each DPT, we follow the design of DPT-hybrid~\cite{ranftl2021dpt} by using $M=6$ transformer layers for encoding and $M=6$ for decoding. In our case, we use an input resolution of 256$\times$320 and patch size $p=16$. A ResNet-50~\cite{he2016deep} acts as the patch embedding backbone. For output head design, we take output features from stage 1 and 2 of ResNet, as well as output from layer 3 and 6 from decoder, fuse the representations and use 4 convolutions layers with 2 bilinear interpolation layers to produce the final output.  Readout token~\cite{ranftl2021dpt} is set to \textit{ignore} and batch normalization (BN) is enabled in output heads. The only difference among the DPTs is the output layer in the head, depending on the sub-task. We use \textit{tanh} activation for all heads to output albedo, roughness, normals and inverse depth.
Details on the tensor sizes and head design are in the supplementary material.

For \textbf{LightNet} we have a similar encoder-decoder design. Three independent heads are required for estimating axis center, intensity, and bandwidth of $K$ Spherical Gaussians for each pixel. However we found that sharing decoders for the three tasks is not optimal as the output spaces of these tasks are very disparate, thus forcing a unified decoder feature space destabilizes training. As a result, we use a shared encoder but independent decoders and output heads for \textbf{LightNet}. We use 4 transformer layers in both encoders and decoders in the multi-task setting so that the entire model can fit into one GPU for joint training. We use a similar output layer design as aforementioned except for not using \textit{tanh} for $\xi_k$ but use normalization to unit norm instead.

\subsection{Multi-task Network Design}

The single-network design requires 4 DPTs for material and geometry and one for lighting. As a result, the collective memory footprint is too large to fit the entire model into a regular GPU for training. An alternative option is to allow DPTs to have a unified feature space so that memory usage can be reduced. Also in some cases, a jointly learned feature can benefit from related tasks. Inspired by UniT~\cite{hu2021unit} where input-domain-specific encoders and shared decoders are designed for a multi-modal input setting, we use a unified encoder and decoder for all tasks in \textbf{BRDFGeoNet} besides independent task-specific convolutional heads. We share decoders due to memory considerations while noting that further gains from independent decoders might be possible.
The result is $\mathcal{B}_{multi}$ with 4 heads as shown in Fig.~\ref{fig:pipeline}.
The design for \textbf{LightNet} is the same as in a single-task setting.

\subsection{Additional Components}
\noindent{\textbf{Refinement using bilateral solver.}} 
We may optionally refine the geometry and material outputs with a bilateral solver (BS). Additional refinement leads to smoother outputs, which is preferable for some metrics like WHDR~\cite{bell2014iiw} for albedo. In comparison to previous CNN-based works, we observe that our transformer-based outputs are already quite accurate without the bilateral solver on all the tasks.


\vspace{0.2cm}
\noindent{\textbf{Cascade design.}} In prior works~\cite{li2020inverse}, a cascade design is used to refine the predictions based on the rendering error. However, it leads to a two-fold increase in memory with small improvements, while we already achieve significantly improved results on all benchmarks with a single-stage network. As a result, we choose not to use cascaded refinement.

\begin{figure*}
\centering
\includegraphics[width=0.95\textwidth]{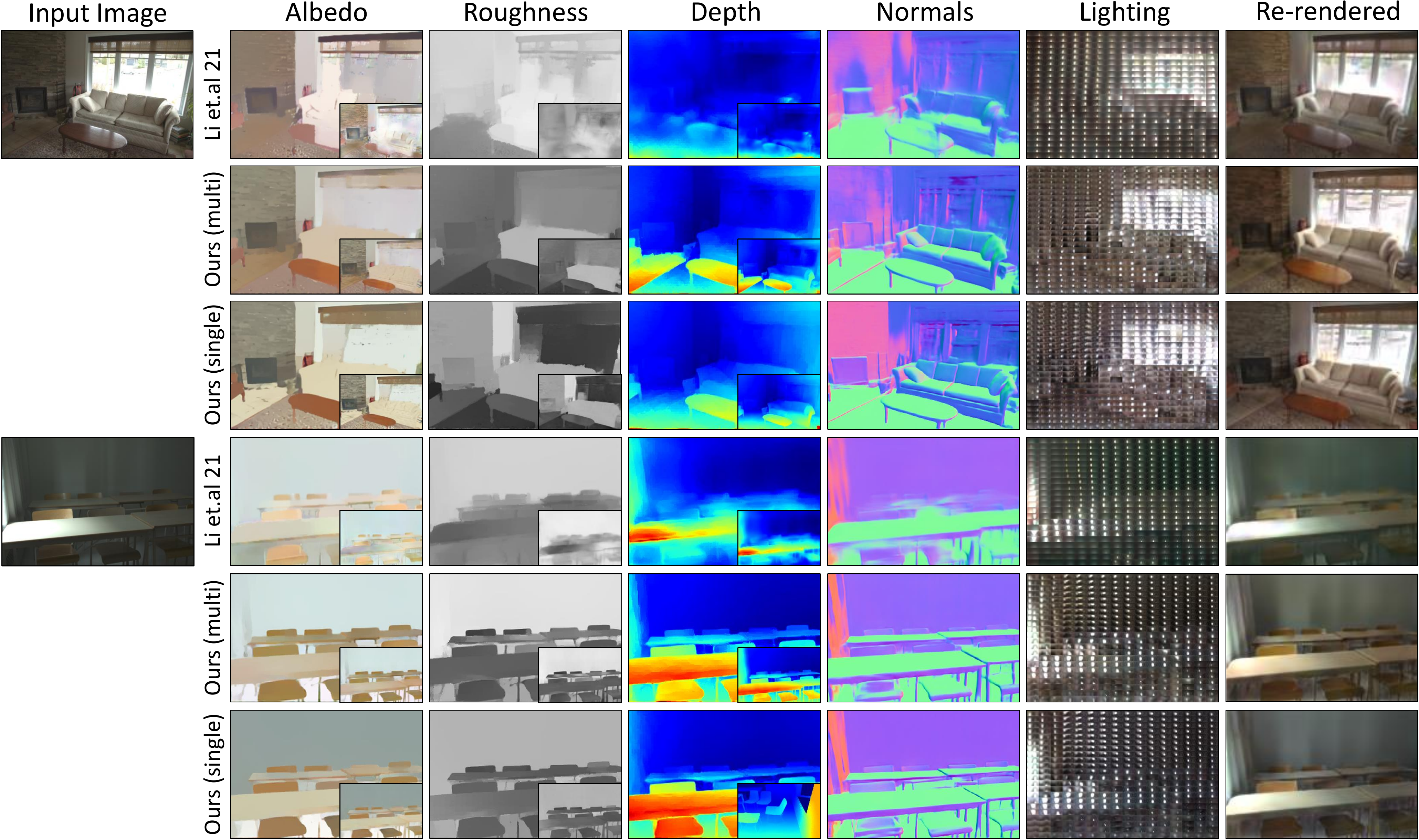}
\vspace{-0.3cm}
\caption{BRDF, geometry estimation, per-pixel lighting and re-rendering results on Garon~\etal~\cite{garon2019fast} (after BS). Insets are results before BS. Our material and geometry results are cleaner even without BS. Also less artifacts can be observed in our re-rendered images (see bright area on the table in sample 2) compared to Li~\etal~\cite{li2021openrooms} due to better estimation on all modalities.}
\label{fig:real_results}
\vspace{-0.4cm}
\end{figure*}

\section{Evaluation}
\label{sec:eval}
\vspace{-0.2cm}



We demonstrate the capability of our DPT-based \Ours{} to produce globally coherent estimations that outperform traditional CNN-based models on all modalities, due to its global attention that can better handle the inherent ambiguities of inverse rendering. This is especially notable for material and lighting prediction, in the presence of highlights, shadows, and interreflections. We include results on joint BRDF, geometry, and lighting prediction on OpenRooms, as well as sub-tasks on real world benchmarks. We additionally provide analysis on design choices and ablation study.

\subsection{Datasets and Training}
\vspace{-0.2cm}

Given the success of synthetic inverse rendering datasets in providing photorealistic images and complete ground truth for all inverse rendering tasks, we use OpenRooms (OR)~\cite{li2021openrooms} dataset for supervised training. We use 6,684 scenes for training, 1,008 for testing, each rendered with multiple material and lighting configurations, for a total of 102,452 images for training, and 15,738 frames for testing. 


We train \textbf{BRDFGeoNet} (in multi-task and single-task settings) on OR with Adam optimizer for 80 epochs at a learning rate of 1e-5 and batch size of 8 on 4 GPUs, starting with pretrained ResNet on ImageNet. Then we freeze \textbf{BRDFGeoNet} and train \textbf{LightNet} in the same setting. Additional training details and weights for losses are in the supplementary material.

After training on OpenRooms, we may finetune on real datasets where labels of all or a subset of the tasks are available. Specifically, we finetune on (a) IIW dataset~\cite{bell2014iiw} with relative labels of albedo; (b) NYUv2~\cite{silberman2012nyu} with ground truth depth and normals. We also demonstrate lighting estimation results with virtual object insertion on real world images from Garon~ \etal~\cite{garon2019fast} where ground truth lighting is collected at selected locations with light probes.


\begin{figure*}
  \begin{minipage}[c]{0.70\textwidth}
    \includegraphics[width=\textwidth]{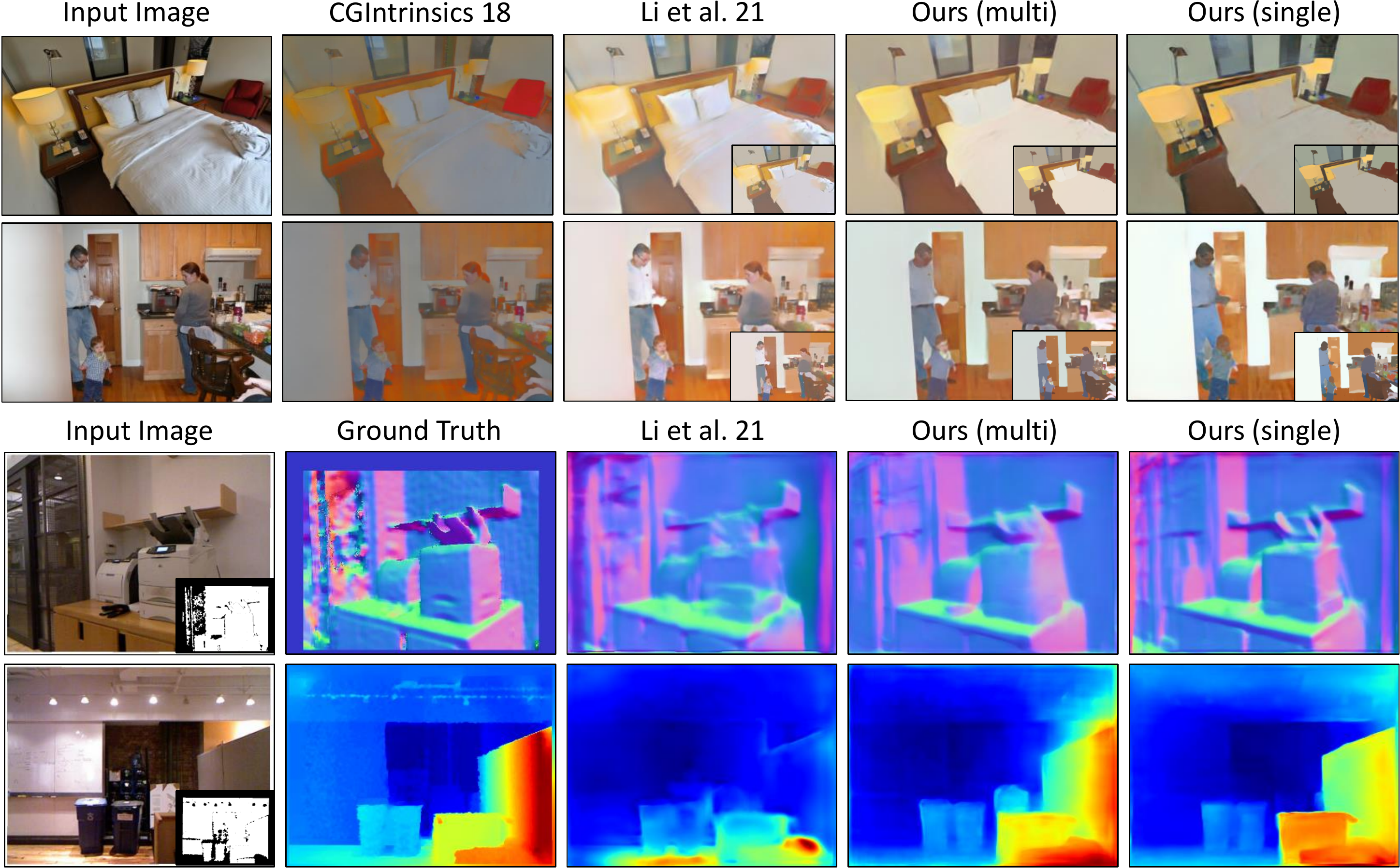}
  \end{minipage}\hfill
  \begin{minipage}[c]{0.28\textwidth}
    \caption{(Top) Intrinsic decomposition results on IIW~\cite{bell2014iiw} (before BS). The insets are results after BS. Our results are better in regions with complex geometry and lighting (see the bedding in sample 1 and clothing in sample 2). (Bottom) Geometry estimation results on NYUv2~\cite{silberman2012nyu} (all without BS). Ours are less prone to artifacts (see printer surface in sample 1 and geometry of the trash bins in sample 2). Please refer to the supplementary material for more results.
    } \label{fig:iiw_nyu_results}
  \end{minipage}
\end{figure*}

\subsection{BRDF and Geometry Estimation}

Table~\ref{tab:brdfgeo} includes the performance of \Ours{} as well as Li~\etal~\cite{li2021openrooms} (which we trained and finetuned in the same setting as ours for all evaluations) for BRDF and geometry estimation, evaluated on our OpenRooms test split, both with a variant where the bilateral solver is applied to albedo, roughness, and depth. We observe better performance from both our multi-task and single-task models consistently on all tasks, and we visually demonstrate the comparison with a few samples in Fig.~\ref{fig:or_results}. As can be observed, our model excels with much cleaner and accurate material and geometry estimations, and better lighting estimations (especially in areas of highlights and shadows where we better disentangle lighting from albedo and correspondingly yields brighter/darker lighting).

\begin{table}[h]
  \small
  \centering
  \setlength{\tabcolsep}{0.3em}
    \begin{tabular}{c c c c c c c c}
    \toprule
    {Method} & \textbf{A}$\downarrow$ & \textbf{R}$\downarrow$ & \textbf{D}$\downarrow$ & \textbf{N}$\downarrow$ & \textbf{L}$\downarrow$ & \textbf{I}$\downarrow$ & \textbf{L}+\textbf{I}$\downarrow$\\
    \hline
    {\Ours{}~(multi)}& 0.51 & 5.52 & 1.72 & 2.05 & 12.50 & 1.15 & 12.54\\
    {\Ours{}~(multi+BS)} & 0.51 & \underline{5.50} & 1.71 & 2.05 & 12.47 & 1.15 & 12.58\\
    {\Ours{}~(single)}& \textbf{0.43} & \underline{5.50} & \textbf{1.42} & \textbf{1.89} & \textbf{12.04} & 0.99 & \textbf{12.14}\\
    {\Ours{}~(single+BS)}& \textbf{0.43} & \textbf{5.48} & \underline{1.44} & \textbf{1.89} & \underline{12.08} & 0.97 & \underline{12.17}\\
    {Ours (direct)}& - & - & - & - & 12.29 & 1.29 & 12.42\\
    Li'21~\cite{li2021openrooms} & 0.52 & 6.31 & 2.20 & 2.61 & 18.63 & \textbf{0.88} & 18.72\\
    Li'21+BS~\cite{li2021openrooms} & 0.48 & 6.30 & 1.91 & 2.61 & 18.61 & \textbf{0.88} & 18.70\\
    \bottomrule
    \end{tabular}%
    \caption{Errors of BRDF, geometry and lighting with a base of $10^{-2}$ on OpenRooms~\cite{li2021openrooms}. Lower is better. For lighting estimation, \textbf{L} is the lighting reconstruction error, \textbf{I} is the rendering error and \textbf{L}+\textbf{I} is the combined lighting loss for which \textbf{LightNet} is trained.}
    \label{tab:brdfgeo}
\end{table}

\subsection{Lighting Estimation}
 In Table~\ref{tab:brdfgeo}, we report lighting estimation errors of different versions of \Ours{}, including both multi-task and single-task models, and variants of these two models with BS. We also include another version of our lighting estimation model (denoted as \textit{Ours (direct)}), where the first stage of BRDF and geometry estimation is removed, and only the image is directly used as input for lighting prediction. We observe the best lighting prediction from the single models due to their larger capacity. For the direct model, it is able to estimate reasonable lighting directly from image input, highlighting the power of the transformer architecture, but is not suitable for downstream tasks (\eg object insertion, material editing) where a complete scene decomposition is required.
 

\subsection{Comparison on Sub-tasks}

\noindent{\textbf{Intrinsic decomposition.}}
To evaluate \Ours{} on the task of intrinsic decomposition (albedo-only) on real world images, we finetune our model on IIW~\cite{bell2014iiw} using relative labels of albedo between labeled pairs of pixels. Results are summarized in Table~\ref{tab:iiw} and Fig.~\ref{fig:iiw_nyu_results}. We observe that our single-task model performs better than the multi-task version due to its larger capacity and independent feature space. More importantly, both the multi-task and single-task models achieve new state-of-the-art on IIW, outperforming all prior methods. Note that the shift in tones is due to the weak supervision from relative loss.

\begin{table}[!!t]
\small
  \centering
  \setlength{\tabcolsep}{0.3em}
    \begin{tabular}{c c c}
    \toprule
    \multicolumn{1}{c }{Method}& Finetuning Datasets & WHDR$\downarrow$ \\
    \hline
    \multicolumn{1}{c }{\Ours{}~(multi)}& OR+IIW & \underline{13.1}\\
    \multicolumn{1}{c }{\Ours{}~(single)}& OR+IIW & \textbf{12.0}\\
    Li'21~\cite{li2021openrooms} & OR+IIW & 16.4 \\
    Li'20~\cite{li2020inverse} & CGM+IIW & 15.9 \\
    NIR'19~\cite{sengupta2019nir} & CGP+IIW & 16.8 \\
    CGIntrinsics'18~\cite{li2018cgintrinsics} & CGI+IIW & 17.5 \\
    \bottomrule
    \end{tabular}%
    \vspace{-0.2cm}
    \caption{Intrinsic decomposition on IIW~\cite{bell2014iiw}. Lower is better.}
    \label{tab:iiw}
    \vspace{-0.4cm}
\end{table}%

\begin{table}[!!t]
\small
  \centering
  \setlength{\tabcolsep}{0.3em}
    \begin{tabular}{c c c c}
    \toprule
    \multicolumn{1}{c }{Method}& Mean(\degree)$\downarrow$ & Med.(\degree)$\downarrow$ & Depth$\downarrow$\\
    \hline
    \multicolumn{1}{c }{\Ours{}~(multi)}& 23.5 & 16.3 & \underline{0.162} \\
    \multicolumn{1}{c }{\Ours{}~(single)}& \textbf{20.2} & \textbf{13.4} & \textbf{0.132}\\
    Li'21~\cite{li2021openrooms} & 25.3 & 18.0 & 0.171 \\
    Li'20~\cite{li2020inverse} & 24.1 & 17.3 & 0.184 \\
    NIR'19~\cite{sengupta2019nir} & \underline{21.1} & 16.9 & - \\
    Zhang'17~\cite{zhang2017physically} & 21.7 & \underline{14.8} & - \\
    \bottomrule
    \end{tabular}%
    \vspace{-0.2cm}
    \caption{Normal (mean and median) and depth (mean on inverse depth) prediction results on NYUv2~\cite{silberman2012nyu}. Lower is better.}
    \vspace{-0.2cm}
    \label{tab:geo}
\end{table}%


\noindent{\textbf{Geometry estimation.}}
For depth and normal prediction, we follow the training and evaluation settings of Li~\etal~\cite{li2021openrooms}, and report results in Table~\ref{tab:geo} and Fig.~\ref{fig:iiw_nyu_results}. We choose to compare with similar methods in a multi-task inverse rendering setting, instead of dedicated and more complex methods that maximize geometry prediction performance in the wild like DPT~\cite{ranftl2021dpt} or MiDaS~\cite{ranftl2019midas}. As can be observed, we achieve improved results compared to all previous works listed. 

\begin{figure}
\centering
    \includegraphics[width=\columnwidth]{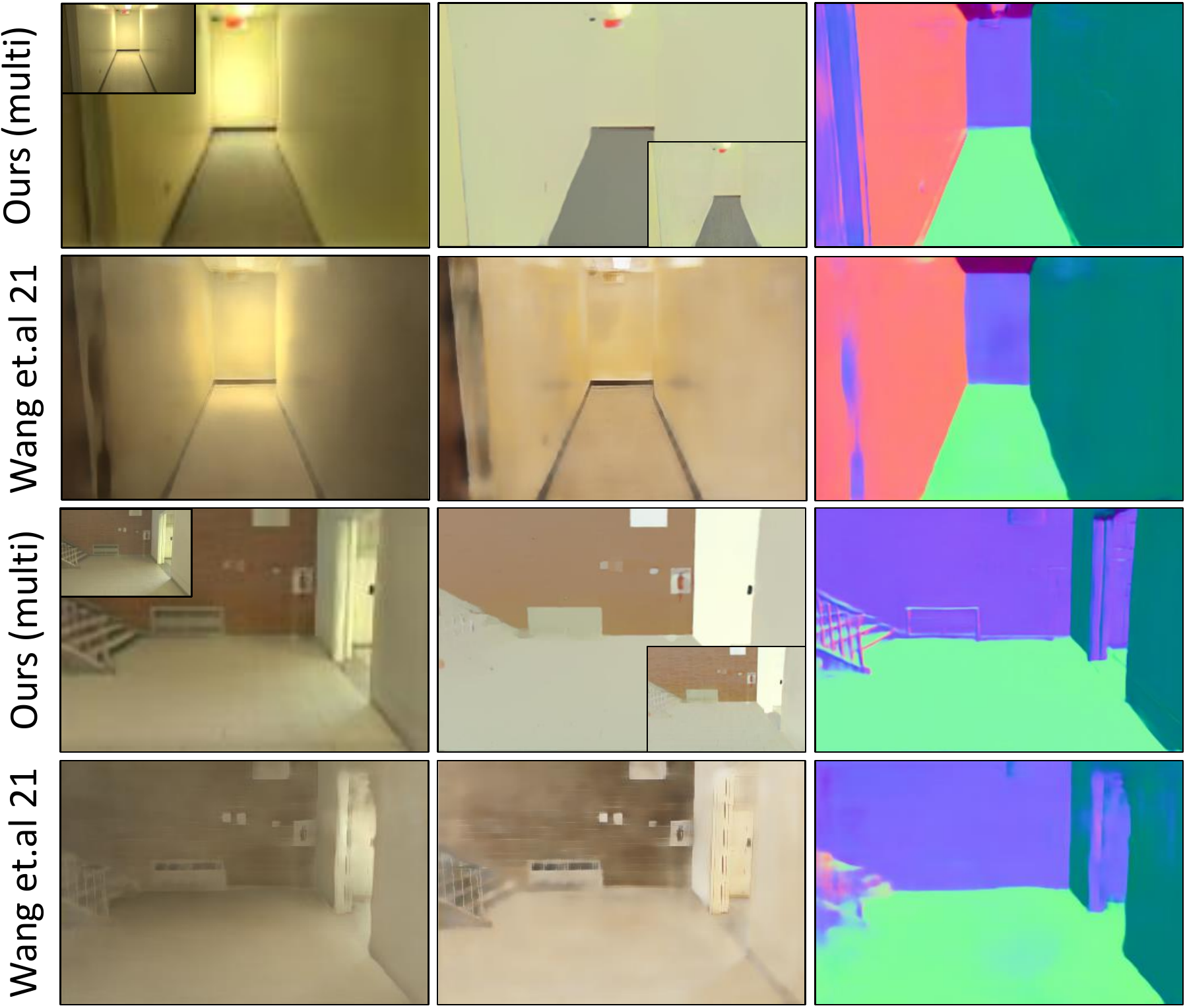}
    \vspace{-0.4cm}
    \caption{Comparison between our results and Wang~\etal~\cite{wang2021learning} on albedo, normals and re-rendering. On challenging inputs, we achieve smoother albedo with less artifacts and richer details with more consistent normals. Upper-left insets are input images, and lower-right insets are before BS).
    } \label{fig:ot_compare}
  \vspace{-0.2cm}
\end{figure}

\vspace{0.1cm}
\noindent{\textbf{Intermediate results on real world images.}}
In Fig.~\ref{fig:real_results} we test \Ours{} on real world images from Garon~\etal~\cite{garon2019fast}, and we demonstrate that \Ours{} generalize well to real world images and outperform previous art in every task. In general, our results are more spatially consistent and have fewer artifacts (most noticeable from the re-rendered images which are the result of all estimations). In Fig.~\ref{fig:ot_compare}, we compare with the recent method of Wang~\etal~\cite{wang2021learning} on their reported samples, where we arrive at much improved results.

\begin{table}[!!t]
\small
  \centering
  \setlength{\tabcolsep}{0.3em}
    \begin{tabular}{c c c c}
    \toprule
    Gardner'17~\cite{gardner2017learning} & Garon'19~\cite{garon2019fast} & Li'21~\cite{li2021openrooms} & Ground Truth\\
    \hline
    0.24 & 0.30 & 0.47 & 0.58 \\
    \bottomrule
    \end{tabular}%
    \vspace{-0.2cm}
    \caption{A user study on object insertion, where we compare \Ours{} with each of the previous work or ground truth and report the percentage of feedbacks where other method is considered to be more photorealistic than ours.}
    \vspace{-0.2cm}
    \label{tab:user}
\end{table}%

\begin{table}[!!t]
\small
  \centering
  \setlength{\tabcolsep}{0.3em}
    \begin{tabular}{c c c c c}
    \toprule
    \multicolumn{1}{c }{}& single-6 & single-4 & multi & Li'20~\cite{li2021openrooms}\\
    \hline
    \multicolumn{1}{c }{Model Size (MB)}& 7,305 & 6,256 & 1,539 & 795 \\
    \multicolumn{1}{c }{Inference (ms)}& 141.9 & 125.9 & 91.9 & 45.2\\
    \multicolumn{1}{c }{\textbf{A}+\textbf{R}+\textbf{D}+\textbf{N}}& 6.00 & 6.08 & 6.44 & 7.65 \\
    \multicolumn{1}{c }{\textbf{L}+\textbf{I}}& 12.14 & 12.85 & 12.54 & 18.72 \\
    \bottomrule
    \end{tabular}%
    \vspace{-0.2cm}
    \caption{Analysis on multiple design choices: \Ours{} (single-task with 6 or 4 layers in \textbf{BRDFGeoNet}, multi-task with 4 layers), and CNN-based architecture from Li~\etal~\cite{li2021openrooms} on OR~\cite{li2021openrooms}.}
    \vspace{-0.4cm}
    \label{tab:choices}
\end{table}%

\begin{figure*}
\centering
\includegraphics[width=0.90\textwidth]{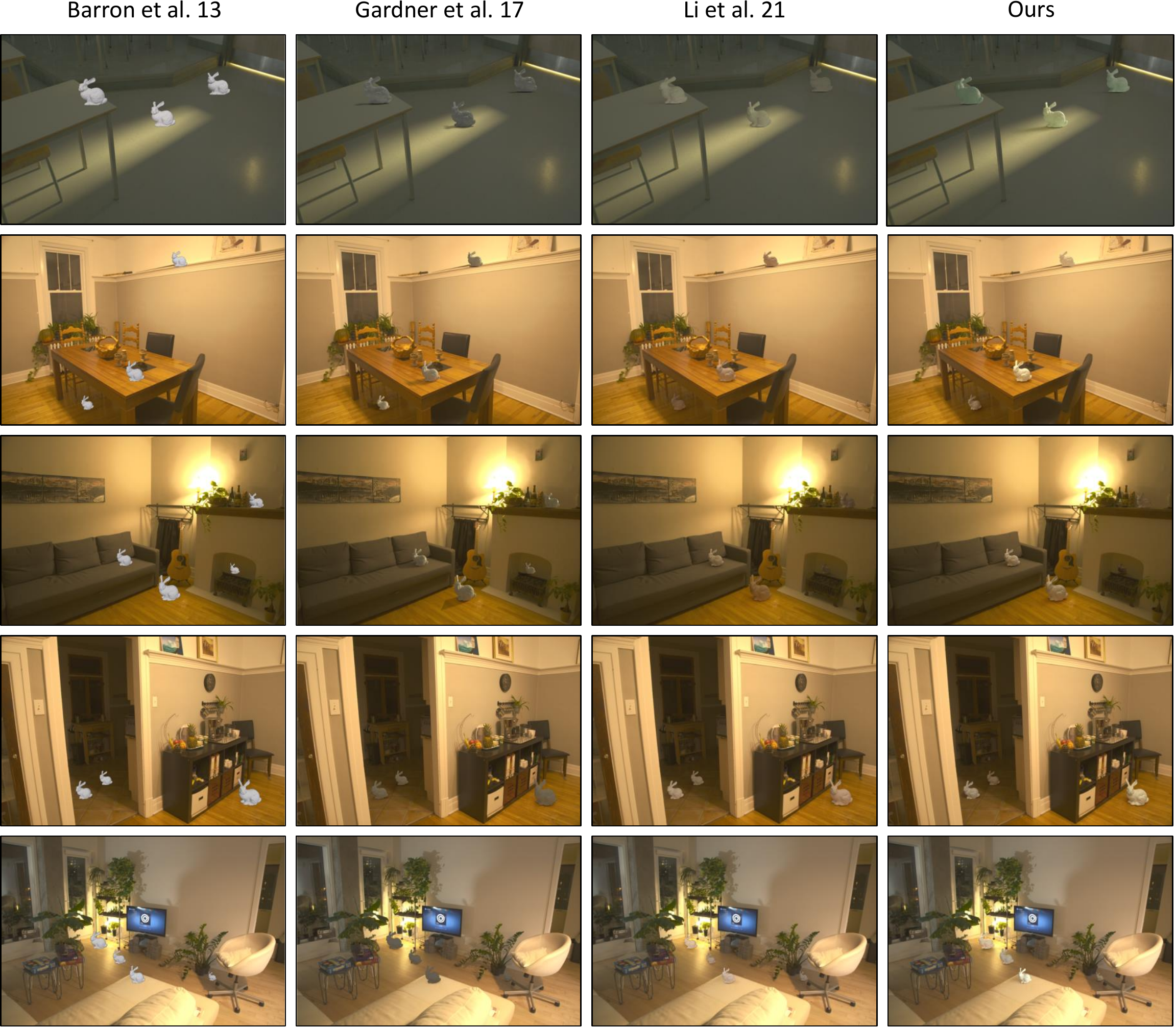}
\vspace{-0.2cm}
\caption{Virtual object insertion results. Lighting estimation from Barron~\etal~\cite{barron2013intrinsic} lacks high frequency, while Gardner~\etal~\cite{gardner2017learning} predicts a global environment map instead of spatially-varying lighting at each location. Compared to the most recent Li~\etal~\cite{li2021openrooms}, we better decouple lighting and appearance to better recover highlighted or shadowed areas (see center object in sample 1 and 2, right object in sample 4 and 5). Also our lighting is more spatially consistent \wrt the light source (see shadow directions in objects of sample 1 and 3).}
\label{fig:insertion_results}
\vspace{-0.4cm}
\end{figure*}

\vspace{0.1cm}
\noindent{\textbf{Lighting estimation on real images.}}
With intermediate estimations including material, geometry, and the final per-pixel lighting, we demonstrate applications of \Ours{} in downstream applications including virtual object insertion and material editing. For object insertion, we compare with prior works in Fig.~\ref{fig:insertion_results}. We produce more realistic lighting for inserted objects which better match the surroundings on lighting intensity, direction and relative brightness of inserted objects in highlights and shadowed areas. 

\begin{figure}[!!t]
\centering
\includegraphics[width=\columnwidth]{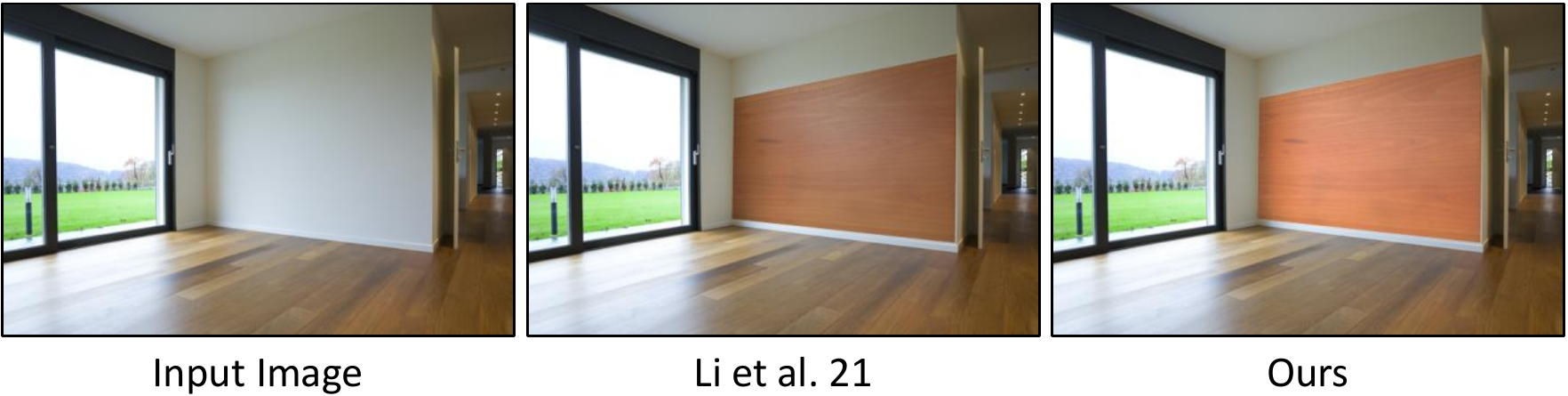}
\vspace{-0.4cm}
\caption{Material editing example where we replace the material of part of the wall with wood. Note that the shadow from outside lighting is recreated on the replaced material, demonstrating our accurate  spatially-varying lighting estimation.}
\label{fig:material}
\vspace{-0.2cm}
\end{figure}

To quantitatively evaluate insertion results, we conduct a user study against other methods in Tab.~\ref{tab:user}. We outperform all previous methods, only being inferior to ground truth. We also perform material editing in Fig.~\ref{fig:material} where we replace the material of a planar surface and re-render the region, to showcase that \Ours{} captures the directional lighting effect of the area so that the replaced material will be properly shadowed. A complete list of results and comparisons is in the supplementary material. 

\subsection{Ablation study}
\label{ablation}
\vspace{-0.2cm}

\begin{figure}[!!t]
  \begin{minipage}[c]{0.6\columnwidth}
    \includegraphics[width=\columnwidth]{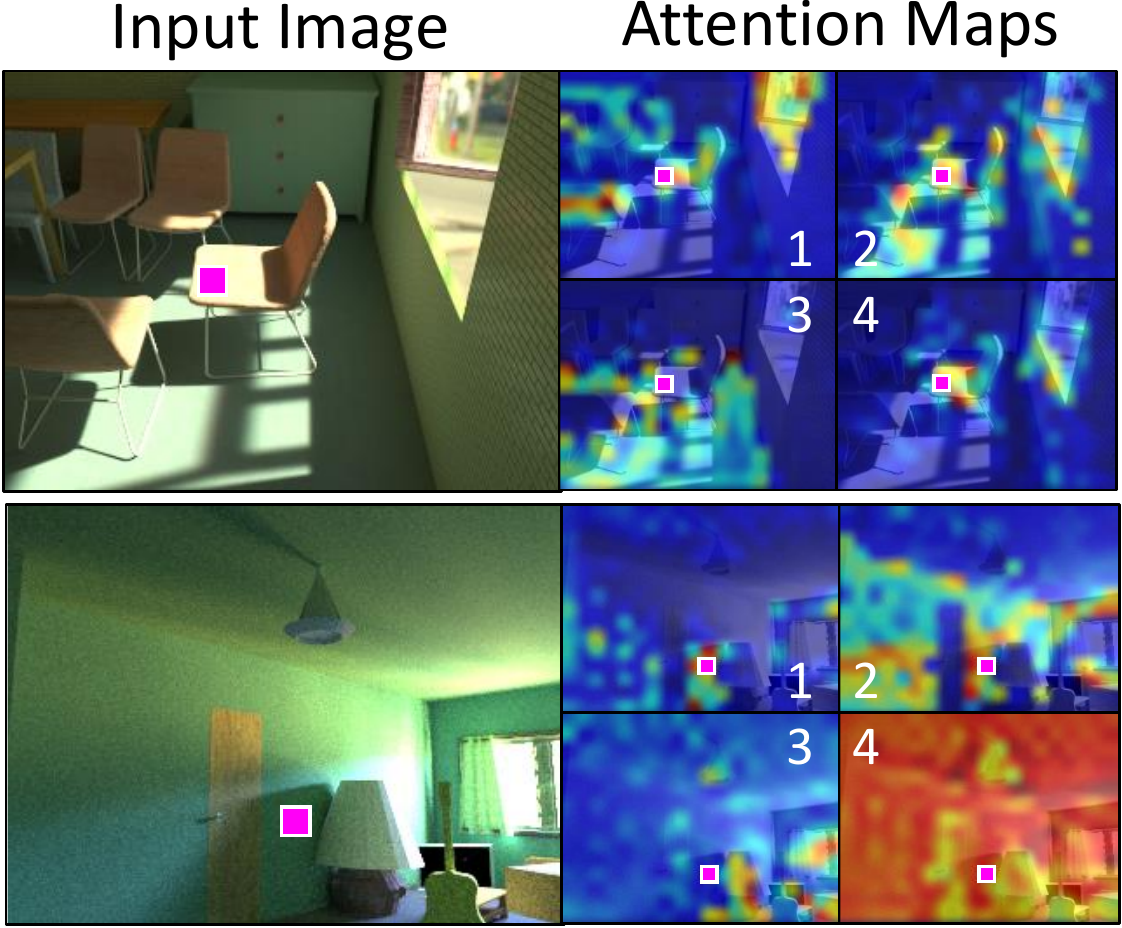}
  \end{minipage}\hfill
  \raisebox{-0.3cm}{
  \begin{minipage}[c]{0.38\columnwidth}
    \caption{Attention maps learned by the single-task model for albedo. Each heatmap is the attention  weights (affinity) of the patch (denoted by pink square) \wrt all other patches, of one head from subsequent transformer layers.
    } 
    \label{fig:attention}
  \end{minipage}}
  \vspace{-0.4cm}
\end{figure}

\noindent{\textbf{Comparison of design choices.}}
In Table~\ref{tab:choices} we compare various metrics of all models, including model size, inference time for one sample on a Titan RTX 2080Ti GPU, test losses on OR for material and geometry combined, and lighting losses combined. Study on other minor choices 
can be found in the supplementary material.

\vspace{0.1cm}
\noindent{\textbf{Attention from Transformers.}}
To provide additional insight into the attention that is learned, we include in Fig.~\ref{fig:attention} two samples where attention maps corresponding to one patch location from different layers are visualized. In the first sample we show a patch on the lit-up chair seat, subsequently attending to (1) chairs and window, (2) highlighted regions over the image, (3) entire floor, (4) the chair itself. For the second sample, the chosen patch is in the shadow on the wall, and it attends to (1) neighbouring shadowed areas of the wall, (2) the entire wall, (3) potential light sources and occluders, (4) ambient environment. Throughout the cascaded transformer layers, \Ours{} learns to attend to large regions and distant interactions to improve its prediction in the presence of complex light transport effects.


\section{Discussion}
\label{sec:conclusion}
\vspace{-0.1cm}

\noindent{\textbf{Limitation and potential negative impact.}}
\Ours{} only infers per-pixel lighting on the scene surface, so applications like inserting objects in the air are not feasible. 
Future work may also explore choices beyond the current multi-task design, possibly by leveraging the complementary nature of various tasks. Potential negative impacts  include Deepfake~\cite{westerlund2019deepfake}, where our method can be used to recreate an indoor scene with a photorealistically modified appearance.

\vspace{0.1cm}
\noindent{\textbf{Conclusion.}}
  We have proposed an inverse rendering framework that estimates material, geometry, and per-pixel lighting given an unconstrained indoor image using a transformer-based model. Our results demonstrate that the model can produce significantly better results especially on material and lighting, which require long-range reasoning for diambiguation. Additionally, our approach enables different design choices with single or multi-task settings. Downstream applications including object insertion and material editing on real world images demonstrate the strength of our model to better handle challenging lighting conditions and produce highly photorealistic results. We also provide analysis into design choices and the attention maps learned by our model.
  

\small
\noindent\textbf{Acknowledgments: }
We thank NSF CAREER 1751365, NSF IIS 2110409 and NSF CHASE-CI, generous support by Qualcomm, as well as gifts from Adobe and a Google Research Award.
\normalsize

\appendix

\section{Summary of Supplementary Material}
The supplementary material is organized as follows:
\begin{tight_itemize}
\item More results from evaluation. (Sec.~\ref{supp:more_results})
\item Details and ablation study on model design. (Sec.~\ref{supp:model_design})
\item Details on the training scheme. (Sec.~\ref{supp:training})
\item Specifications on the user study. (Sec.~\ref{supp:user_study})
\item Discussion on potential negative social impacts. (Sec.~\ref{supp:discussion})
\end{tight_itemize}

\section{More Results}\label{supp:more_results}
\subsection{OpenRooms}

\begin{figure*}
\centering
\includegraphics[width=0.95\textwidth]{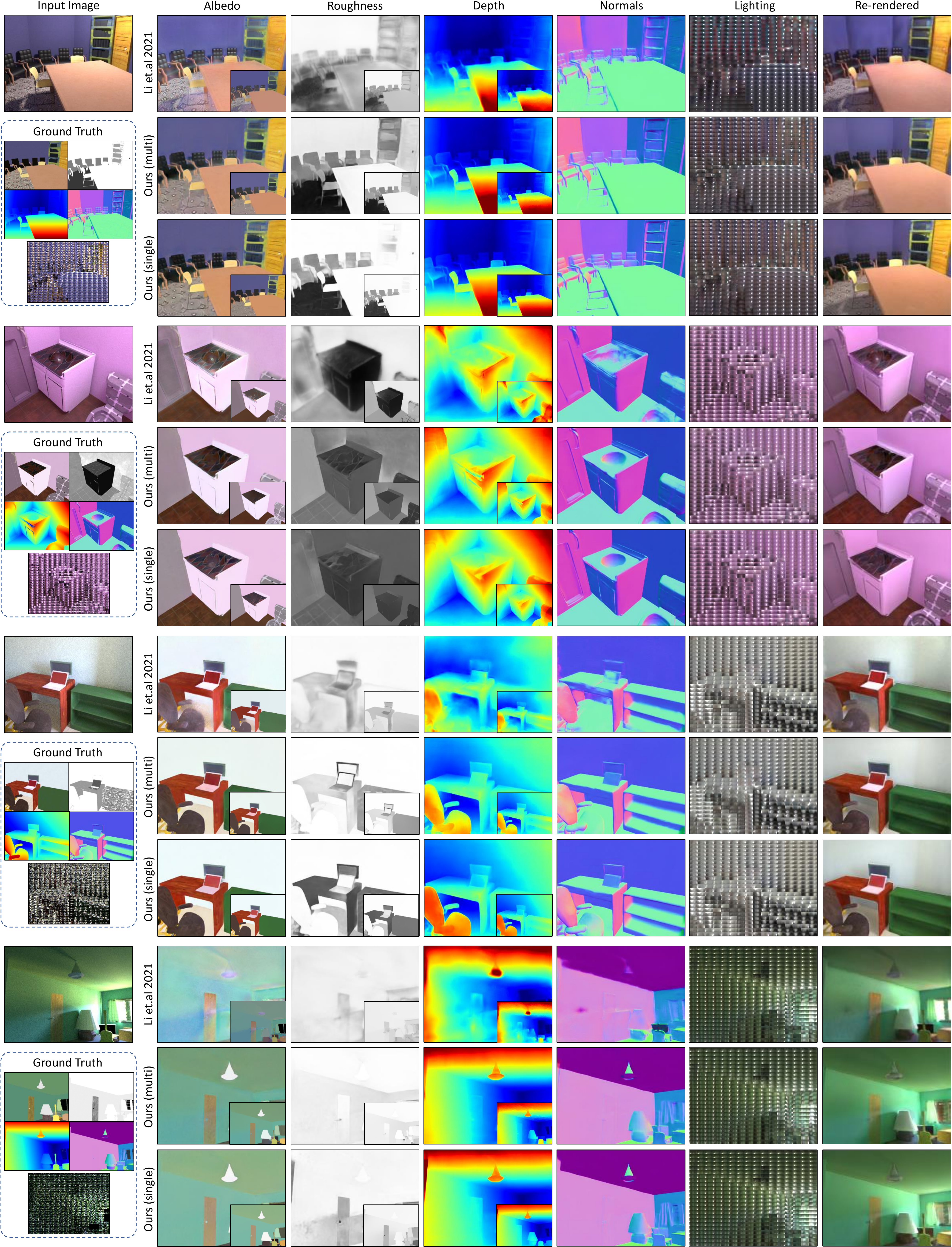}
\vspace{-0.2cm}
\caption{Additional BRDF, geometry and lighting estimation on OpenRooms. Small insets (best viewed when enlarged in PDF version) are estimations processed with bilateral solvers (BS).}
\vspace{-0.4cm}
\label{fig:or_results_more}
\end{figure*}

We include more examples of material, geometry and lighting estimation on OpenRooms in Fig.~\ref{fig:or_results_more}. In general we demonstrate more spatially consistent material estimation (\eg the textured floor in bright and shadowed areas in sample 1, or the surface of the sink table in sample 2), better geometry in challenging lighting (\eg the shape of the sink bowl in sample 2, consistency of the flat wall in samples 3 and 4), as well as fewer artifacts in re-rendered results as a result of overall better estimation of all factors.

\subsection{Real World Images}

\begin{figure*}
\centering
\includegraphics[width=\textwidth]{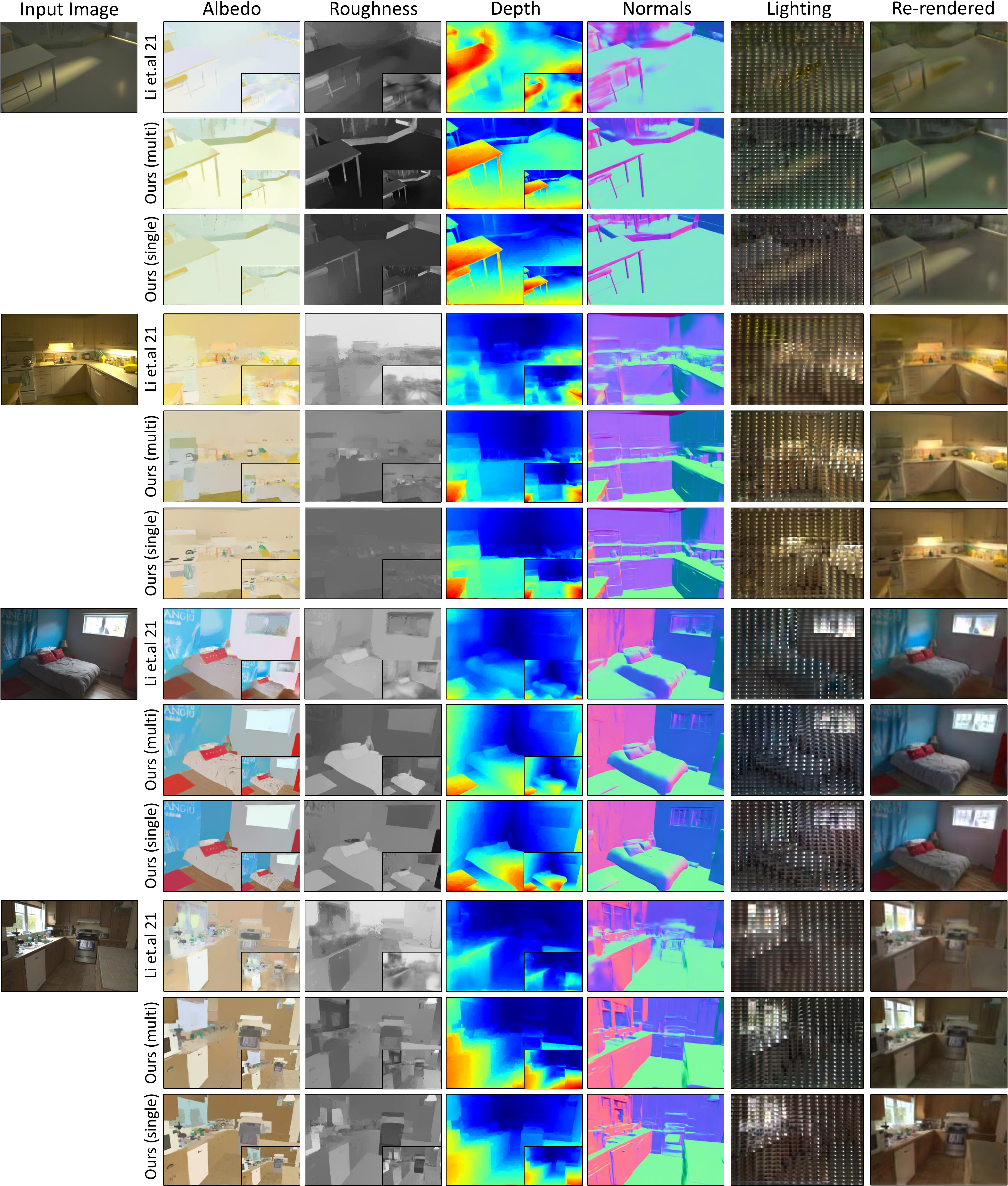}
\vspace{-0.3cm}
\caption{Additional BRDF, geometry estimation, per-pixel lighting and re-rendering results on Garon~\etal~\cite{garon2019fast} (after BS). Insets are results before BS.}
\label{fig:real_results_more}
\vspace{-0.4cm}
\end{figure*}

We include more examples of material, geometry and lighting estimation on real world images from Garon~\etal~\cite{garon2019fast} in Fig.~\ref{fig:real_results_more}. We arrive at similar observations on comparisons of geometry, material and lighting as in the previous subsection, and we prove that our methods generalize well to real world indoor images in the wild.

\subsection{IIW}

\begin{figure*}
\centering
\includegraphics[width=\textwidth]{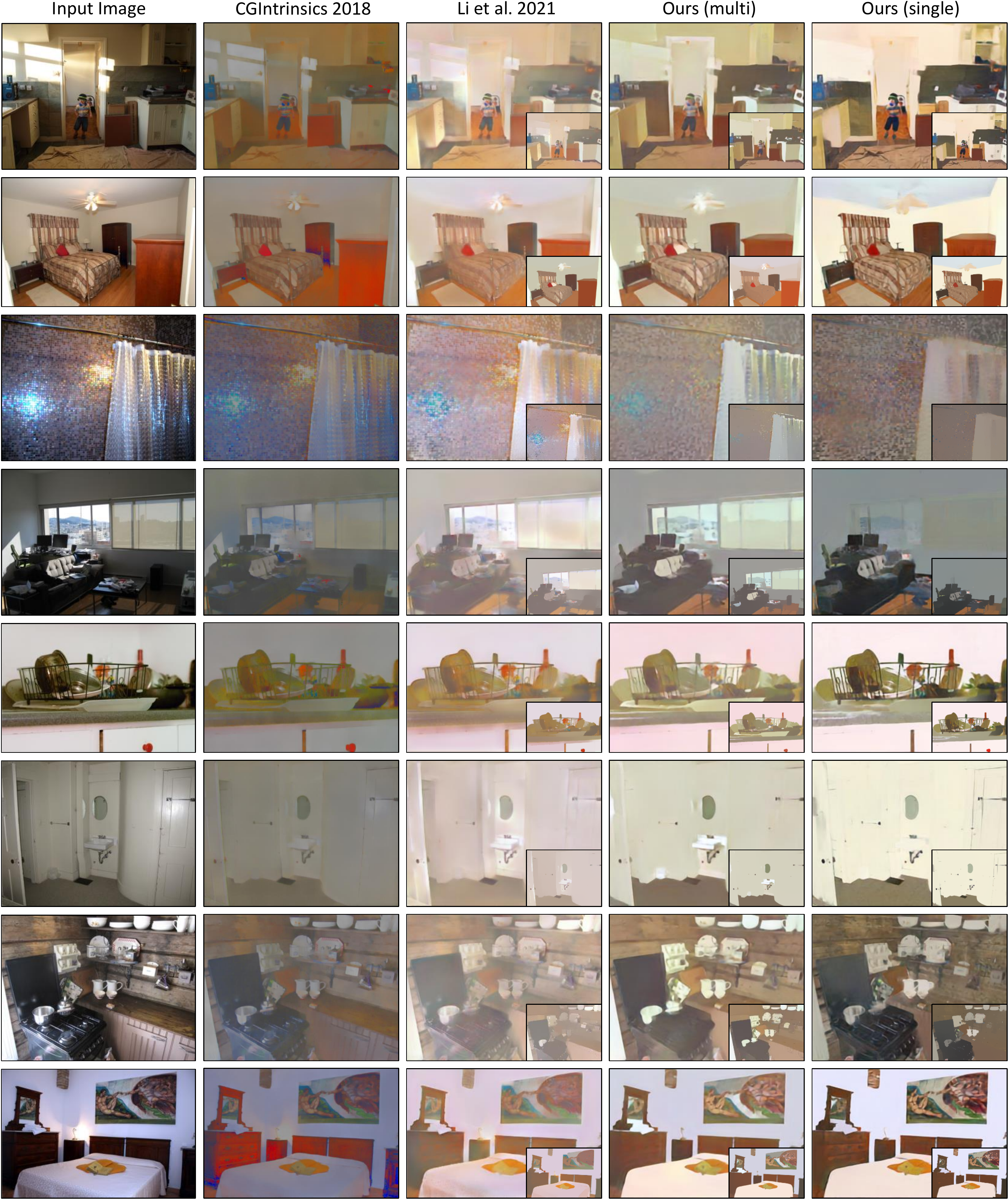}
\caption{Additional intrinsic decomposition results on IIW~\cite{bell2014iiw} (all before BS). The inset figure within each result is the result after BS.}
\label{fig:iiw_results_more}
\end{figure*}

We include more examples of albedo estimation on IIW in Fig.~\ref{fig:iiw_results_more} to showcase our state-of-the-art albedo estimation from our models finetuned on IIW, with better spatial consistency and rich details.

\subsection{NYUv2}

\begin{figure*}
\centering
\includegraphics[width=\textwidth]{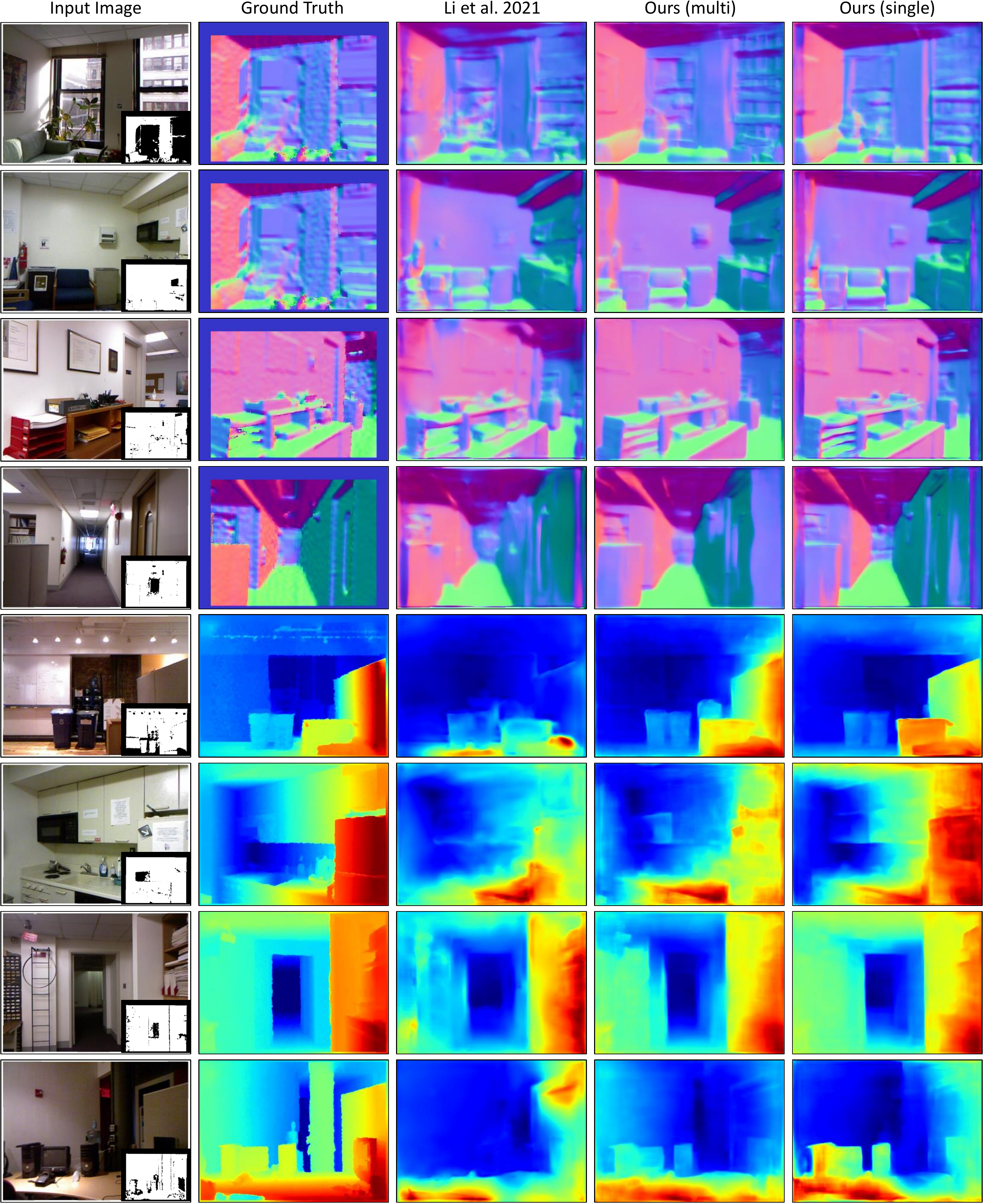}
\caption{Additional geometry estimation results on NYUv2~\cite{silberman2012nyu} (all without BS).}
\label{fig:nyu_results_more}
\end{figure*}

We include more examples of depth and normal estimation on NYUv2 in Fig.~\ref{fig:nyu_results_more} to show improved geometry estimation compared to previous works in multi-task inverse rendering setting.

\subsection{Object Insertion}

\begin{figure*}
\centering
\includegraphics[width=0.95\textwidth]{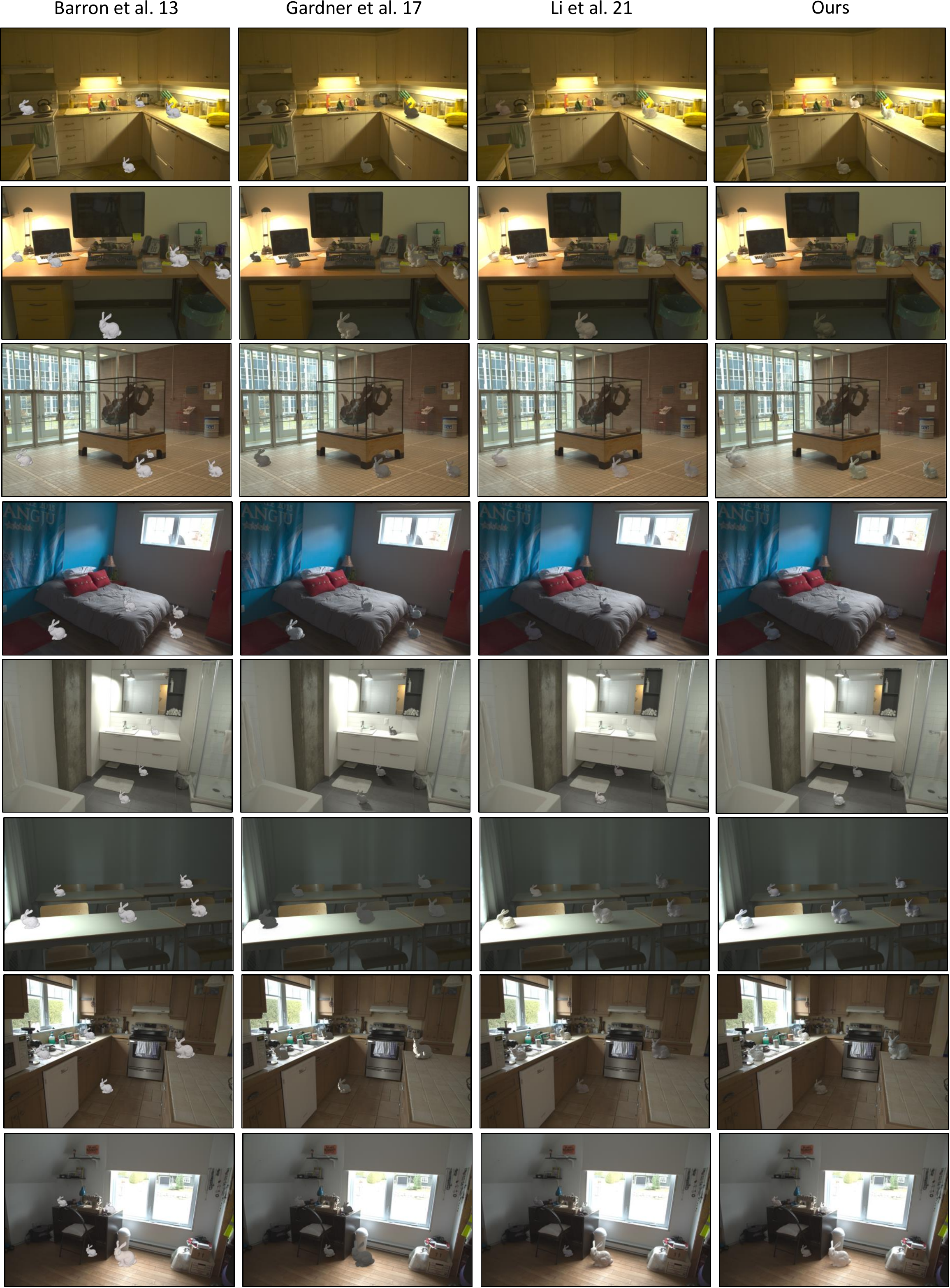}
\vspace{-0.2cm}
\caption{Additional virtual object insertion results.}
\label{fig:insertion_results_more}
\vspace{-0.4cm}
\end{figure*}

We show in Fig.~\ref{fig:insertion_results_more} more samples of virtual object insertion where we achieve with more photorealistic results. In particular, we show more physically plausible lighting with better spatial consistency (\eg lighting on the bunnies which sit against major light sources to the camera in sample 2 and 4), strong and accurate directional lighting (\eg the bunnies around the lighting sources in sample 1 and 2 where they are properly lit up or darkened according to their relative position and orientation \wrt the kitchen/desk lamps).

\subsection{Material Editing}
\begin{figure}
\centering
\includegraphics[width=\columnwidth]{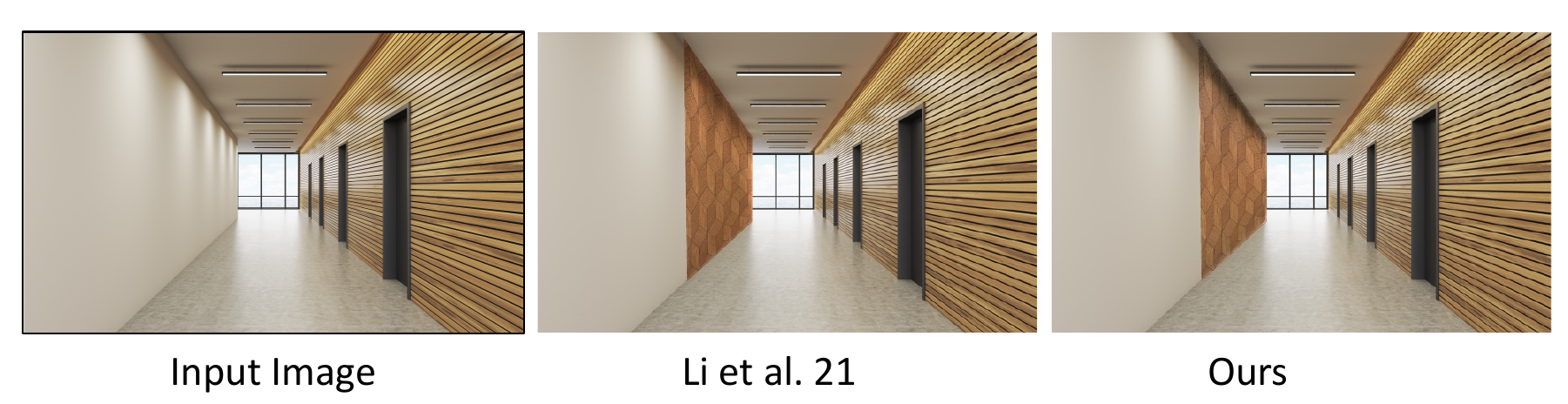}
\vspace{-0.2cm}
\caption{An additional material editing result.}
\vspace{-0.4cm}
\label{fig:matEdit}
\end{figure}

We show in Fig.~\ref{fig:matEdit} an additional material editing result. We observe that our method can recover spatially-varying lighting, with the rendered result similar to that of prior state-of-the-art \cite{li2020inverse}. 

\subsection{Attention Maps}
\begin{figure*}[!!t]
\centering
\includegraphics[width=\textwidth]{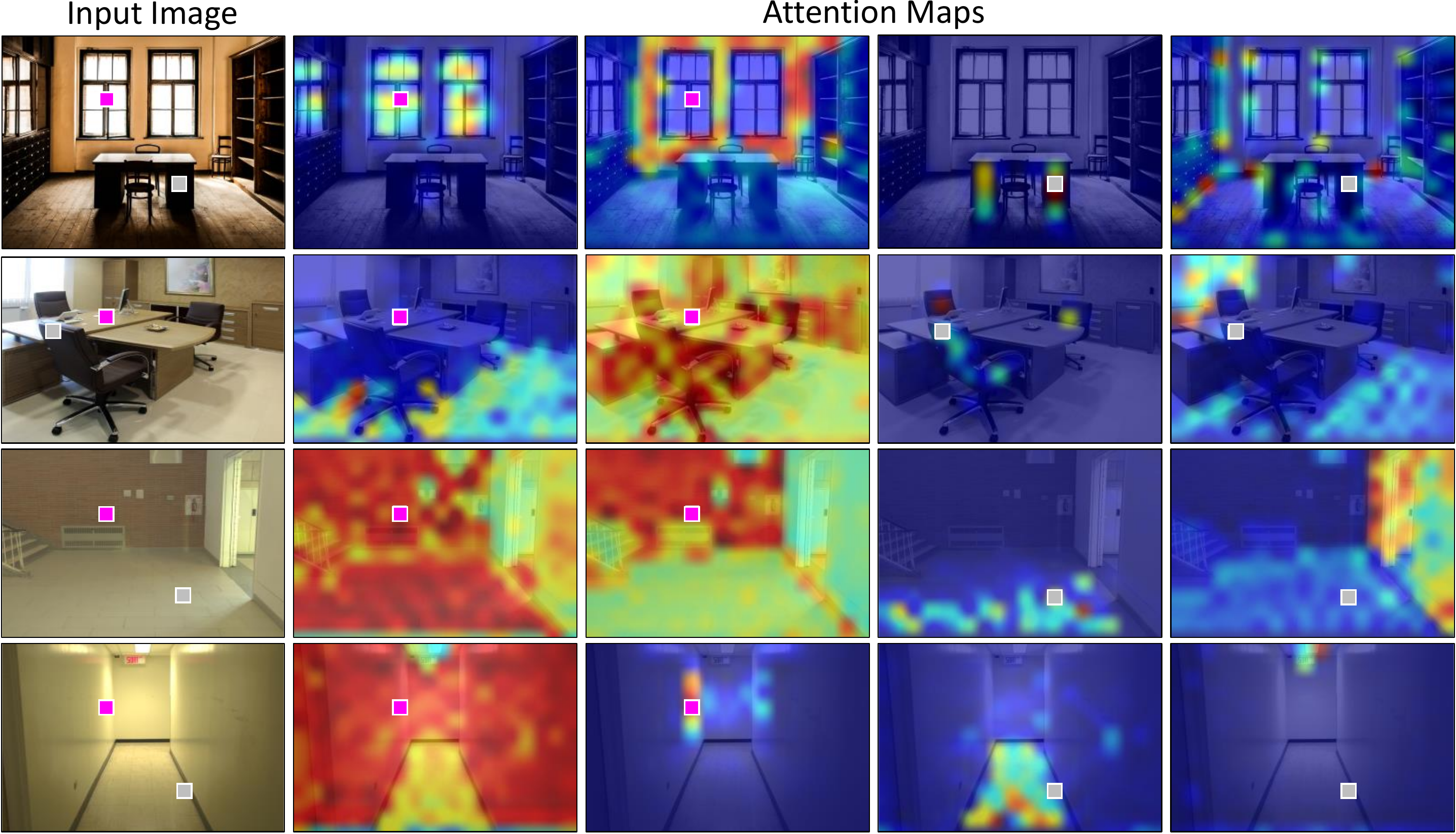}
\caption{More attention maps learned by the single-task model for albedo, on real world images. We pick 4 samples of real world images, and for each image show two attention maps for each of two selected token locations.} 
\label{fig:attention_more}
\vspace{-0.2cm}
\end{figure*}

We include more examples of the attention maps of selected patches on real world images in Fig.~\ref{fig:attention_more}. Our model learns to attend to various semantic regions within the image (\eg the entire object, other objects, area of highlights or shadows, \etc) to update the token (feature) for a patch, without explicit supervision of semantic regions. This attention across potentially long-range interactions results in better disambiguation of the shape, material and lighting factors.

\section{Model Design Details}\label{supp:model_design}

\subsection{Detailed Architecture}

\begin{figure*}
\centering
\includegraphics[width=\textwidth]{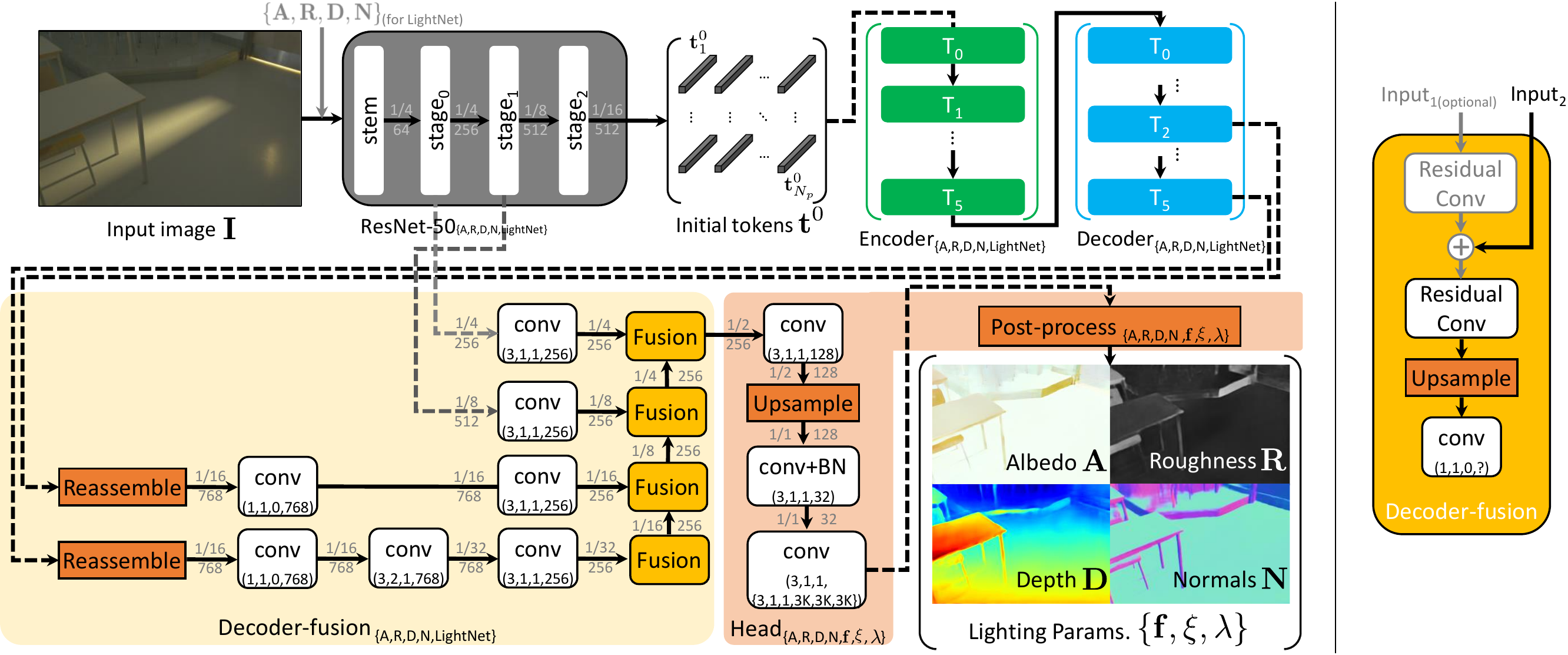}
\vspace{-0.2cm}
\caption{Details of the modules of the single-task and multi-task models (left) and the Decoder-fusion module (right). \texttt{Reassemble} operation is introduced in DPT~\cite{ranftl2021dpt}. Details of the \texttt{Post-process} operation for different modalities can be found in text. The ratio and integer in gray font around each arrow are respectively the output tensor size \wrt original input size to the network, and output tensor channels.}
\vspace{-0.4cm}
\label{fig:pipeline_detailed}
\end{figure*}

The modules in our model mostly follow the design of DPT-hybrid~\cite{ranftl2021dpt} as detailed in Fig.~\ref{fig:pipeline_detailed}. We denote the convolutional operation as \texttt{conv(k,s,p,c)} where \texttt{k} is the convolutional kernel size, \texttt{s} is stride, \texttt{p} is padding, and \texttt{c} is output channels. \texttt{BN} is for batch normalization, \texttt{upsample} is for 2x bilinear interpolation. All convolution layers are followed with \textit{ReLU} activation unless otherwise stated, and with Batch Normalization except in \texttt{Head} where BN is only optionally applied to the second convolution layer (further comparison included in Subsection~\ref{sec:BN}). We use the same residual attention module in Transformer layers as in DPT~\cite{sengupta2019nir} or ViT~\cite{dosovitskiy2020vit} where each layer is followed by \textit{GELU} activation~\cite{hendrycks2016gelu} and Layer Normalization~\cite{ba2016layern}.

\noindent\textbf{Head design.}
We use BN in all heads of \textbf{BRDFGeoNet} as we found it to be useful to stabilize training. However we choose not to use BN in \textbf{LightNet} heads as it will tend to force the model to converge to a local minimum which produces blurry results. The choice of BN is mostly empirical but we include additional comparison on BN in Table~\ref{tab:bn} where we disable BN for \textbf{BRDFGeoNet} or enable BN for \textbf{LightNet} and compare with previous results to support our choice.

All heads share similar architecture as illustrated in Fig.~\ref{fig:pipeline_detailed} except for the \texttt{Post-process} layer where non-trainable normalization or activation operations are applied depending on the specific domain. For albedo, normals and roughness we use \textit{tanh} activation so that the output is constrained in a closed range of [-1, 1] due to the physical nature of those properties. Albedo and roughness prediction are later re-scaled to [0, 1] while normals are normalized to have unit L2 norm. For depth, we also use \textit{tanh} activation plus scaling to predict inverse depth within (0, 1] and then invert to linear depth space.

For \textbf{LightNet} heads of bandwidth of SGs $\{\lambda_k\}_{k=1}^K$ and intensities $\{\textbf{f}_k\}_{k=1}^K$ we again use \textit{tanh} and re-scale output to [0, 1]. For estimating axes $\{\xi_k\}_{k=1}^K$, we use the same rule as estimating normals via \textit{tanh} plus normalization.

\subsection{Ablation Study on Model Design}
We include comparison on several design choices and report the performance. The comparisons include (1) sharing or not sharing decoders in \textbf{BRDFGeoNet} in multi-task setting, where we provide comparison on joint estimation of albedo+roughness (material tasks) in Table~\ref{tab:al+ro_share} or albedo+normal (material and geometry jointly estimated) in Table~\ref{tab:al+no_share}, considering the full multi-task model with 4 independent decoders does not fit into our hardware for training; (2) BN or no BN, where we compare in single-task estimation of depth, albedo and lighting in Table~\ref{tab:bn}; (3) 4 layers in encoder/decoder vs. 6 layers, in single-task models in Table~\ref{tab:4vs6} as an addition to Table 5 in the main paper. We conclude that, (1) there is not a significant performance drop by sharing decoders, or using 4 layer in Transformers, but there are benefits for significant memory saving; (2) BN works better for \textbf{BRDFGeoNet} heads while no BN is preferred for \textbf{LightNet} heads. We demonstrate that, by empirically comparing those choices we arrive at our final architecture, which achieves reasonable trade-off between memory cost and accuracy.
\label{sec:BN}

\begin{table}[!!t]
\small
  \centering
  \setlength{\tabcolsep}{0.3em}
    \begin{tabular}{c c c}
    \toprule
    \multicolumn{1}{c }{al+ro}& sharing decoders & not sharing decoders\\
    \hline
    \multicolumn{1}{c }{Model Size (MB)}& 1,206 & 1.606\\
    \multicolumn{1}{c }{Inference (ms)}& 34.4 & 42.3\\
    \multicolumn{1}{c }{$L_\mathbf{A}$}& \textbf{0.50} & 0.51\\
    \multicolumn{1}{c }{$L_\mathbf{R}$}& 1.91 & \textbf{1.88}\\
    \bottomrule
    \end{tabular}%
    \vspace{-0.2cm}
    \caption{Analysis on whether to share decoders in multi-task joint estimation of albedo and roughness: comparison on model sizes, inference speed and losses.}
    \vspace{-0.4cm}
    \label{tab:al+ro_share}
\end{table}%

\begin{table}[!!t]
\small
  \centering
  \setlength{\tabcolsep}{0.3em}
    \begin{tabular}{c c c}
    \toprule
    \multicolumn{1}{c }{al+no}& sharing decoders & not sharing decoders\\
    \hline
    \multicolumn{1}{c }{Model Size (MB)}& 1,206 & 1.606\\
    \multicolumn{1}{c }{Inference (ms)}& 34.4 & 42.3\\
    \multicolumn{1}{c }{$L_\mathbf{A}$}& \textbf{0.51} & \textbf{0.51}\\
    \multicolumn{1}{c }{$L_\mathbf{N}$}& 1.88 & \textbf{1.85}\\
    \bottomrule
    \end{tabular}%
    \vspace{-0.2cm}
    \caption{Analysis on whether to share decoders in multi-task joint estimation of albedo and normals: comparison on model sizes, inference speed and losses.}
    \vspace{-0.4cm}
    \label{tab:al+no_share}
\end{table}%

\begin{table}[!!t]
\small
  \centering
  \setlength{\tabcolsep}{0.3em}
    \begin{tabular}{c c c}
    \toprule
    \multicolumn{1}{c }{}& BN & no BN\\
    \hline
    \multicolumn{1}{c }{$L_\mathbf{A}$ in single-task albedo estimation}& \textbf{0.43} & 0.51\\
    \multicolumn{1}{c }{$L_\mathbf{N}$ in single-task normal estimation}& \textbf{1.89} & 1.92\\
    \multicolumn{1}{c }{$L_\mathbf{light}$ in multi-task lighting estimation}& 13.23 & \textbf{12.54}\\
    \bottomrule
    \end{tabular}%
    \vspace{-0.2cm}
    \caption{Analysis on whether to use BN in output heads in single-task estimation of albedo and normals, as well as multi-task estimation of lighting: comparison on losses.}
    \vspace{-0.4cm}
    \label{tab:bn}
\end{table}%

\begin{table}[!!t]
\small
  \centering
  \setlength{\tabcolsep}{0.3em}
    \begin{tabular}{c c c}
    \toprule
    \multicolumn{1}{c }{}& single-4 & single-6\\
    \hline
    \multicolumn{1}{c }{$L_\mathbf{A}$}& 0.48 & \textbf{0.43}\\
    \multicolumn{1}{c }{$L_\mathbf{R}$}& 1.93 & \textbf{1.91}\\
    \multicolumn{1}{c }{$L_\mathbf{D}$}& 1.43 & \textbf{1.42}\\
    \multicolumn{1}{c }{$L_\mathbf{N}$}& \textbf{1.89} & \textbf{1.89}\\
    \bottomrule
    \end{tabular}%
    \vspace{-0.2cm}
    \caption{Analysis on 4-layer encoder-decoder design vs. 6-layer versions in single-task estimation of albedo, roughness, normals and depth, as well as multi-task estimation of lighting: comparison on losses.}
    \vspace{-0.4cm}
    \label{tab:4vs6}
\end{table}%
\section{Training Details}\label{supp:training}
We train our entire pipeline in two stages: (1) train \textbf{BRDFGeoNet} with full supervision on albedo, roughness, depth and normals, (2) freeze \textbf{BRDFGeoNet}, feed the output to \textbf{LightNet} and train \textbf{LightNet} with full supervision on the estimated lighting map and re-rendered image. We additionally use binary masks on pixels of objects $\mathbf{M}_o \in \mathbb{R}^{h\times w}$ or $\mathbf{M}'_o \in \mathbb{R}^{h\times w\times 3}$ (excluding windows), and masks on pixels of valid materials and lighting $\mathbf{M}_l \in \mathbb{R}^{h\times w}$ or $\mathbf{M}'_l \in \mathbb{R}^{h\times w\times 3}$ (excluding surface of lit-up lamps and windows).

For the first stage, as stated in Sec.3.1, we use scale-invariant L2 loss~\cite{li2020inverse, ranftl2019midas} for albedo and depth ($\log$ space) and L2 loss for roughness and normals. Specifically, for losses on roughness and normals, we have:

\begin{equation}
    L_{\mathbf{R}} = ||(\mathbf{R} - \mathbf{\hat{R}}) \cdot \mathbf{M}_l||^2_2, 
\end{equation}

\begin{equation}
    L_{\mathbf{N}} = ||(\mathbf{N} - \mathbf{\hat{N}}) \cdot \mathbf{M}'_o||^2_2.
\end{equation}

For albedo, the loss is

\begin{equation}
    L_{\mathbf{A}} = ||(\mathbf{A} - \mathbf{\hat{A}'}) \cdot \mathbf{M}'_l||^2_2,
\end{equation}

\noindent where $\mathbf{\hat{A}'}$ is the estimated albedo aligned to the ground truth from a  least-squares solution~\cite{li2020inverse}. For depth, the loss is computed in $\log$ space:

\begin{equation}
    L_{\mathbf{D}} = ||(\log\mathbf{D} - \log\mathbf{\hat{D}'}) \cdot \mathbf{M}'_o||^2_2
\end{equation}

\noindent where similarly $\mathbf{\hat{D}'}$ is the estimated depth aligned in $\log$ space to the ground truth from a least-squares solution~\cite{li2020inverse}.

Given we train the entire pipeline in two stages, we break $L_{\textrm{all}}$ into two losses; $L_{\textrm{all}} = L_{\textrm{BRDFGeo}} + L_{\textrm{light}}$. In training \textbf{BRDFGeoNet} in multi-task setting, the loss $L_{\textrm{BRDFGeo}}$ is a weighted combination of losses on albedo, roughness, depth and normals:

\begin{equation}
    L_{\textrm{BRDFGeo}} = \lambda_{\textbf{A}} L_{\textbf{A}} + \lambda_{\textbf{R}} L_{\textbf{R}} + \lambda_{\textbf{D}} L_{\textbf{D}} + \lambda_{\textbf{N}} L_{\textbf{N}}, 
    \label{equ:loss_brdfgeo}
\end{equation}

\noindent where $\lambda_{\textbf{A}} = 1.5$, $\lambda_{\textbf{R}} = 0.5$, $\lambda_{\textbf{D}} = 0.5$, $\lambda_{\textbf{N}} = 1.0$.

In single-task settings, the loss for each task will simply be the corresponding loss term weighted by the corresponding weight. In the second stage of training \textbf{LightNet}, the loss is a combination of the lighting map reconstruction loss and the image-space re-rendering loss:

\begin{equation}
    L_{\textrm{light}} = \lambda_{\textbf{L}} L_{\textbf{L}} + \lambda_{\textbf{I}} L_{\textbf{I}}
    \label{equ:loss_light}
\end{equation}

\noindent with $\lambda_{\textbf{L}} = 10.0$ and $\lambda_{\textbf{I}} = 1.0$.

All models are trained with a learning rate of $1e-5$ and a batch size of 8 with Adam optimizer without weight decay, over the entire training set of OpenRooms for 40 epochs until convergence.

\noindent \textbf{Finetuning details.}
For finetuning on IIW and NYUv2, we follow Li~\etal~\cite{li2020inverse} on losses and finetuning strategy for a fair comparison. In each finetuning step, we draw one batch of size 8 from IIW/NYUv2, do a full feed-forward pass and back-propagation with a learning rate of 1e-5, using relative loss on albedo or full supervised loss on normals and depth as described in Li ~\etal~\cite{li2020inverse}. Then we draw another batch from OR of size 8, do the same training step as what is done in pre-training the models on OR. We finetune on IIW/NYU2 for 10 epochs in all finetuning experiments.
\section{User Study Details}\label{supp:user_study}

For the user study, we employ users from Amazon Mechanical Turk to determine the photorealism of an image with inserted bunnies. We compare object insertion results from a set of methods including Gardner'17~\cite{gardner2017learning}, Garon'19~\cite{garon2019fast}, Li'21~\cite{li2021openrooms}, ours and results rendered using ground truth lighting. In each task, we ask the user to do an A/B test where a pair of images from both our method (multi-task setting) and one baseline method are presented. The user is asked to pick the one with `better photorealism' based on how well all the inserted objects blend into the image. Each pair of images is presented to 20 users and for each method we use all 20 images from Garon~\etal~\cite{garon2019fast} with inserted bunnies. We interpret the results as follows: for each comparison of ours against a baseline method, the percentage of users who consider results from the baseline method as better than ours, by averaging 20 feedbacks from each comparison.

\section{Statement on Potential Negative Impacts}\label{supp:discussion}
As stated in the main paper, one possible negative impact of our method is the vulnerability to misuses including Deepfake~\cite{westerlund2019deepfake}. While there is no way to prevent our method to be used by a third-party once it is open-sourced, a way to mitigate the potential negative impact is to employ techniques like Yu~\etal~\cite{yu2021artificial} to embed fingerprinting into the model and hence the results, so that tracing of accountability of improperly edited or generated results will be made possible to countermeasure malicious use. 

Another impact is due to the nature of the transformer architecture we use. Due to the fact that transformers are relatively new and understudied especially for computing efficiency on current dedicated deep learning hardware, as well as their larger computing cost compared to previous CNN-based models on similar tasks, the new models may result in increased carbon footprint if deployed on a large scale. Thus it is important to dedicate more research effort to discover improvements to the architecture and hardware implementation, so that the potential energy and environmental concerns can be mitigated while enabling the use of transformers in inverse rendering.

{\small
\bibliographystyle{ieee_fullname}
\bibliography{egbib}
}

\end{document}